\documentclass[journal]{IEEEtran}

\usepackage[dvipsnames, svgnames, x11names]{xcolor}
\usepackage{amsmath,amssymb}
\usepackage{diagbox}
\usepackage{graphicx}
\usepackage{subfigure}
\usepackage{float}
\usepackage{amsmath}
\usepackage{overpic}
\usepackage{amssymb}
\usepackage{array}
\usepackage{cite}
\usepackage{booktabs}
\usepackage{multirow}

\definecolor{mygreen}{RGB}{32,176,4}
\definecolor{myblue}{RGB}{80,113,193}

\definecolor{new_mypurple}{RGB}{112,48,160}
\definecolor{new_mygreen}{RGB}{0,176,80}

\usepackage{hyperref}
\hypersetup{hidelinks}

\graphicspath{{figures/}}

\newcommand{\etal}{\textit{et al}.}
\newcommand{\eg}{\textit{e}.\textit{g}.}
\newcommand{\ie}{\textit{i}.\textit{e}.}

\begin{document}

\title{Crowd Counting via Perspective-Guided Fractional-Dilation Convolution}


\author{Zhaoyi~Yan, Ruimao Zhang, Hongzhi Zhang,~Qingfu Zhang,~and~Wangmeng~Zuo

\thanks{The submitted manuscript is an extension of our ICCV 2019 paper ``Perspective-guided Convolution Networks for Crowd Counting''. In this version, we have completely rewritten all the paper. Moreover, we present a novel perspective-guided fractional-dilation convolution to allocate spatially variant receptive fields. Additionally, all experimental results are updated and more experiments are conducted for comprehensive evaluation. The conference paper is also provided as the support materials along with the submission.}
\thanks{Z. Yan is with the School of Computer Science and Technology, Harbin Institute of Technology, Harbin 150001, China (e-mail: yanzhaoyi@outlook.com).}
\thanks{R. Zhang is with the School of Data Science, The Chinese University of Hong Kong, Shenzhen, and also with Shenzhen Research Institute of Big Data, Shenzhen 518172, China (e-mail: ruimao.zhang@ieee.org).}
\thanks{H. Zhang is with the School of Computer Science and
Technology, Harbin Institute of Technology, Harbin 150001, China (e-mail: zhanghz@hit.edu.cn).}
\thanks{Q. Zhang is with the Department of Computer Science, City University of Hong Kong, Hong Kong, and also with the Shenzhen Research Institute, City University of Hong Kong, Shenzhen 518057, China. (e-mail: qingfu.zhang@cityu.edu.hk).}
\thanks{W. Zuo is with the School of Computer Science and
Technology, Harbin Institute of Technology, Harbin 150001, China (e-mail: cswmzuo@hit.edu.cn).}
}

\markboth{ieee transactions on multimedia}%
{}



\maketitle
\begin{abstract}


Crowd counting is critical for numerous video surveillance scenarios.
One of the main issues in this task is how to handle the dramatic scale variations of pedestrians caused by the perspective effect.
To address this issue, this paper proposes a novel convolution neural network-based crowd counting method, termed Perspective-guided Fractional-Dilation Network (PFDNet).
By modeling the continuous scale variations, the proposed PFDNet is able to select the proper fractional-dilation kernels for adapting to different spatial locations.
It significantly improves the flexibility of the state-of-the-arts that only consider the discrete representative scales.
In addition, by avoiding the multi-scale or multi-column architecture that used in other methods,  it is computationally
more efficient.
In practice, the proposed PFDNet is constructed by stacking multiple  Perspective-guided  Fractional-Dilation Convolutions (PFC) on a VGG16-BN backbone.
By introducing a novel generalized dilation convolution operation, the PFC can handle fractional dilation ratios in the spatial domain under the guidance of perspective annotations, achieving  continuous scales modeling of pedestrians.
To deal with the problem of unavailable perspective information in some cases, we further introduce an effective perspective estimation branch to the proposed PFDNet, which can be trained in either supervised or weakly-supervised setting once the branch has been pre-trained.
Extensive experiments show that the proposed PFDNet outperforms state-of-the-art methods on ShanghaiTech A, ShanghaiTech B, WorldExpo'10, UCF-QNRF, UCF\_CC\_50 and TRANCOS dataset, achieving MAE $53.8$, $6.5$, $6.8$, $84.3$, $205.8$, and $3.06$ respectively.
%
%
\end{abstract}


%
%

\IEEEpeerreviewmaketitle

\section{Introduction}

\IEEEPARstart{W}{ith} the growth of global population and urbanization, the frequency of crowd gathering consistently rises in recent years. In some scenarios, stampedes and crushes can be life threatening and should always be prevented.
In recent years, with the widespread use of intelligent monitoring~\cite{liu2017provid, wang2017joint, wang2018background} in public areas such as congested squares, stations and markets, it becomes possible to adopt these techniques to manage, control and prevent crowd gathering in cities.
Among various congested scene content analysis techniques, crowd counting~\cite{zhang2016single, sam2017switching, sindagi2017generating, zhang2018sacnn, shen2018crowd}, which aims to estimate the count of the pedestrians in a scene, is one of the fundamental tasks and has drawn considerable research attention.

Accurate counting in a single image remains a challenging topic due to the complex distribution of people, non-uniform illumination, inter- and intra-scene scale variations, cluttering and occlusions, \textit{etc}.
In the past decades, many methods have been proposed, which can be roughly categorised into detection-based~\cite{viola2004robust, dalal2005histograms, wu2005detection, wang2011automatic}, regression-based~\cite{davies1995crowd, he1990texture, lowe1999object, wang2015deep, idrees2013multi,ryan2009crowd, chan2008privacy, chan2009bayesian, chen2012feature, marana1998efficacy, idrees2013multi}, and deep learning-based approaches~\cite{zhang2016single, sam2017switching, sindagi2017generating, zhang2018sacnn, shen2018crowd, sindagi2017generating, zhang2015cross, huang2018body, liu2018leveraging, liu2018decidenet, liu2018crowd, ranjan2018iterative, shi2019revisiting}.
With powerful representation ability of modern convolutional neural networks (CNNs), deep learning-based methods become dominant and have achieved superior performances in terms of accuracy and robustness.

Although significant progress has been achieved in the past decade by using deep models, the dramatic intra-scene scale variations of people due to the perspective effect is still a big problem, limiting the improvement of counting accuracy.
To address this issue,  existing methods~\cite{zhang2016single, sam2017switching,sindagi2017generating, zhang2018sacnn,shen2018crowd} usually adopt multi-scale or multi-column architectures to aggregate the features from different scales.
However, such schemes usually suffer from several limitations.
Firstly, the multi-column architecture~\cite{zhang2016single} inevitably introduces redundant parameters, huge computation burden and difficulty in optimization.
For example, as shown in~\cite{li2018csrnet}, such methods cannot even compete with a deeper CNN due to the high correlation of the features learned by different
columns.
Secondly, they only consider discrete scales,
which is limited when addressing continuous scale variations in practical scenarios.

\begin{figure}[!t]
\centering
\includegraphics[width=0.48\textwidth]{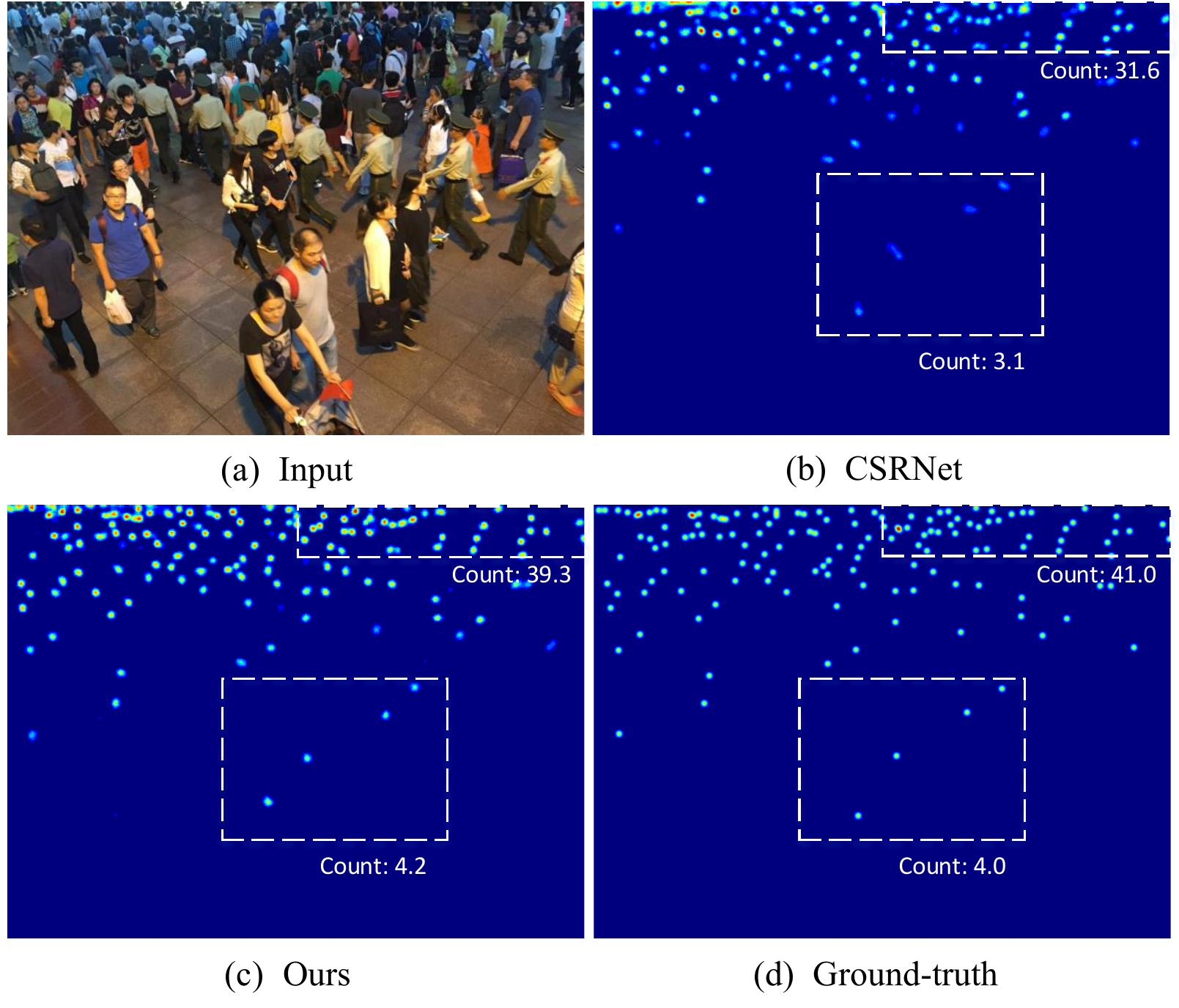}
\scriptsize
\caption{Density maps predicted by CSRNet~\cite{li2018csrnet} and our PFDNet. For this image, the MAE of PFDNet is 3.9, much lower than that of CSRNet (10.1). It is observed that PFDNet delivers consistent superior performances at either smaller or larger scales among the marked regions.}
\label{fig:show_shb}
\end{figure}

Recently, a single-column network architecture termed CSRNet~\cite{li2018csrnet} has been proposed to deal with crowd counting problem.
By applying dilated convolutions, this method achieves  state-of-the-art performance.
However, it still delivers fixed receptive field for different scales
of people, thereby remaining vulnerable to the highly variant intra-scene scales.
As illustrates in Fig.~\ref{fig:show_shb}~(b), this method performs well at intermediate scales, but reveals its incapability for small or large scales of pedestrian heads, indicating its limitation in dealing with scale variations.
Based on the above observation, we argue it is difficult to implicitly model large scale continuous variations by only  using normal components in a multi- or single-column architectures.
Thus we take a step forward to explicitly tackle the scale variations.

%

%
%

Since the perspective information encodes the localized resolution of a scene in the image, it is natural to exploit such annotation as an indication to estimate the scale of pedestrians in a scene.
Thus, this work proposes a novel Perspective-guided Fractional-Dilation Network (\ie, PFDNet), which aims to allocate spatially variant receptive fields under the guidance of perspective information.
In  our implementation,  the  proposed  PFDNet is  constructed  by stacking  multiple Perspective-guided  Fractional-Dilation Convolutions (PFC) on a VGG16-BN  backbone network.
In each PFC, the perspective information is firstly normalized by a customized normalization function and then transformed into a dilation rate map via the linear mapping.
Then we introduce a novel generalized dilation convolution operation, which performs feature aggregation of each position over the sliding window with spatially variant dilation rate, achieving fractional dilation ratios in the spatial domain.
Without bells and whistles, our proposed PFDNet achieves significant performance gain (\eg~$17.4\%$ and ~$33.7\%$ improvement on ShanghaiTech A / B dataset) by replacing normal (dilated) convolutions in CSRNet~\cite{li2018csrnet} with PFC when using ground-truth perspective maps.

In practice, to deal with the problem of unavailable perspective information,
we  further introduce an effective perspective estimation branch to the proposed PFDNet,  which can be trained in either supervised  or weakly-supervised setting once the branch has been pre-trained.
And a three-step estimation strategy is proposed to make the perspective estimation more stable.

In summary, the main contributions of this paper are three folds.
(1) We propose a  Perspective-guided  Fractional-Dilation Convolution (PFC) to mitigate the scale variation in crowd counting.
It is a plug and play module, which is able to work as a spatial allocator of varying receptive field.
(2) Based on PFC, we build up PFDNet for crowd counting, with which a perspective information estimation branch is introduced, making the whole net can be trained with or without perspective annotations.
(3) With the proposed PFDNet, we achieve dramatic performance gain over the state-of-the-arts on several crowd counting benchmarks.
%


This work is an extension of our previous conference paper~\cite{yan2019persp}. Compared with~\cite{yan2019persp}, this paper extends previous method in the following aspects:
\begin{itemize}
\item We propose perspective-guided fractional-dilation convolution, generalizing perspective-guided convolution in~\cite{yan2019persp} .
PFC is a generalized dilation convolution that even works when the dilation rate is a fractional number.
\item PFDNet uses a lightweight single-column architecture compared with PGCNet~\cite{yan2019persp}. While PGC~\cite{yan2019persp} concatenates the smoothed feature maps together with the original feature maps, making the parameters increase dramatically.
\item We also conduct other counting experiments  (\ie~vehicle counting) in this version by using our proposed PFDNet, further validating its broad applications.
\end{itemize}

\section{Related Work}
\label{sec:related_work}

We provide a brief review on the three categories of crowd counting methods, i.e., detection-based, regression-based and CNN-based methods.
Besides, the use of perspective normalization in crowd counting, including PGCNet~\cite{yan2019persp}, is also reviewed.
\subsection{Detection-based Methods}
\label{sec:counting_by_detection}
One straightforward way for single image crowd counting is to conduct human detection and count the number of bounding boxes.
Early researches usually exploit hand-crafted low-level features (\eg, Haar wavelets~\cite{viola2004robust} and histogram of oriented gradient (HOG)~\cite{dalal2005histograms}) which are then fed to the subsequent classifiers.
However, they tend to deliver poor performances when the scene is highly-congested.
To alleviate such issue, part-based detection is proposed for pedestrian counting by detecting particular body parts~\cite{wu2005detection, wang2011automatic}.

\subsection{Regression-based Methods}
\label{sec:counting_by_regression}
Regression-based methods directly learn a mapping from an input image to a numeric crowd count, which gives rise to two other issues, \ie, how to achieve effective feature extraction and how to perform reliable regression.
An early work~\cite{davies1995crowd} adopts holistic features to characterize the global property of a scene, local feature descriptors, \eg, LBP~\cite{he1990texture}, SIFT~\cite{lowe1999object} and even CNN features~\cite{wang2015deep}, have also been introduced to boost the effectiveness of feature extraction~\cite{idrees2013multi,ryan2009crowd}.
In terms of regression models, linear regression and its improvements~\cite{chan2008privacy, chan2009bayesian, chen2012feature, marana1998efficacy} have been widely employed to learn the regression mapping, and in~\cite{idrees2013multi} the multi-scale Markov random field is introduced to aggregate the local counts.

\subsection{CNN-based Methods}
\label{sec:counting_by_cnn}
Many CNN-based  crowd counting methods have been proposed in recent years. These methods usually focus on typical techniques, including multi-scale~\cite{zhang2016single, sam2017switching, sindagi2017generating, zhang2018sacnn, shen2018crowd}, context~\cite{sindagi2017generating}, multi-task~\cite{zhang2015cross, huang2018body, liu2018leveraging}, attention~\cite{liu2019adcrowdnet, sindagi2019ha, zhang2019relational, zhang2019attentional, jiang2020attention}, GNN~\cite{luo2020hybrid}, loss functions~\cite{ma2019bayesian, wang2020DMCount}, classification~\cite{xhp2019SDCNet}, detection~\cite{liu2018decidenet, sam2020locate}, NAS~\cite{hu2020count}, reinforcement learning~\cite{liu2020weighing} and others~\cite{shi2018crowd, liu2018crowd, ranjan2018iterative, wan2019adaptive, sindagi2019multi, shi2019counting, liu2019point, wan2020modeling}.
Recently, some other methods have been proposed in handling the scale variation issue.
For instance, Zhang~\etal~\cite{zhang2016single} suggested a multi-column architecture (MCNN) that combines features with different sizes of receptive fields.
In Switching-CNN~\cite{sam2017switching}, one of the three regressors is assigned for an input image by referring to its specific crowd density.
CP-CNN~\cite{sindagi2017generating} incorporates MCNN with local and global contexts.
SANet~\cite{cao2018scale} employs scale aggregation modules for multi-scale representation.
Instead of multi-column architecture, CSRNet~\cite{li2018csrnet} enlarges receptive fields by stacking dilated convolutions.
Cheng~\etal~\cite{cheng2019learning} aimed to replace Euclidean distance with Maximum Excess over Pixels (MEP) loss and
achieved the promising performance.
Ma~\etal~\cite{ma2019bayesian} proposed Bayesian loss which constructs a density contribution probability model from point annotations.
Liu~\etal~\cite{liu2020adaptive} proposed local counting map and
Wang~\etal~\cite{wang2020DMCount} reformulated crowd counting as a distribution matching problem and propose DM-Count based on optimal transport.

\subsection{Perspective Normalization}
Chan~\etal~\cite{chan2008privacy} first proposed feature normalization for foreground objects under the guidance of perspective information.
Later, Lempitsky~\etal~\cite{lempitsky2010learning} proposed a novel MESA-distance to deal with perspective distortion.
Among the CNN-based methods, a lot of works~\cite{zhang2015cross, zhang2016single, sam2017switching, huang2018body} prefer the perspective information as a prior in generating better density maps, but few of them attempt to encode such information directly into the network architecture.
To our best knowledge, PACNN~\cite{shi2019revisiting} is the most relevant to our method.
However, PACNN is still based on the multi-column architecture, so is limited in only discrete scale representations.
It aims to combine two density maps (predicted by two columns pool4\_2 and pool4\_1 on VGG-16~\cite{simonyan2014very}, respectively) via corresponding weights generated by the perspective map.
In comparison, we argue that a better choice would be to allocate spatially variant receptive fields under the explicit guidance of perspective information during the training process.

\subsection{Perspective-guided convolution}
\label{sec:pgc}
In PGCNet~\cite{yan2019persp}, Yan~\etal~proposed perspective-guided convolution which employs spatially variant Gaussian kernels to smooth the feature maps.
A normal convolution is followed afterwards to aggregate information on the smoothed feature maps.
Generally speaking, we point out that there are three notable differences between PGC~\cite{yan2019persp} and our proposed PFC.

1) The implementation is different. PGC~\cite{yan2019persp} utilizes adaptive Gaussian kernels to perform spatially variant smoothing and obtain the smoothed features, and then aggregate the features via a conventional convolution, while PFC directly samples features by bilinear interpolation on the spatially variant windows.

2) Our PFC is much faster than PGC~\cite{yan2019persp}. Taken feature map with size $256\times 96 \times 128$ as an example, PGC~\cite{yan2019persp} has to generate Gaussian kernel $G_{{\sigma}_{i,j}}$ for each position $(i, j)$ and then performs extra matrix multiplications, resulting in the forward runtime $\sim 10ms$.
In contrast, PFC generalizes the dilated convolution, thus shares similar runtime with conventional dilated convolution (\ie, it takes $\sim 3.3ms$), which only takes $\sim 3.7ms$.
Besides, when the receptive field of allocation rises, due to the intrinsic characteristic of Gaussian smoothing in PGC~\cite{yan2019persp}, the time complexity increases in a quadratic manner attributed to the computation of Gaussian blurring. While, for our proposed PFC, it shares similar runtime with conventional dilated convolution whose runtime nearly keeps static regardless of the changes of dilation rate.

3) PGC~\cite{yan2019persp} suffers from severe feature confusion where our PFC does not suffer.
Due to the spatially variant Gaussian smoothing, the sampling region of Gaussian kernel largely overlaps with each other, which becomes more severe when the size of Gaussian kernel is large.
Such large overlaps inevitably introduce severe mixed information for the smoothed features, which limits the further performance improvement.
While, PFC generalizes dilated convolution where each sampling point in computed via bilinear interpolation of the nearest four grid points, which obviously does not suffer from feature confusion.
%
%

%

\section{Method}
\label{sec:method}
In this section, we first give the definition and implementation details of perspective-guided fractional-dilation convolution (PFC) in Sec~\ref{sec:PGFD}, and then introduce a perspective information estimation branch to the proposed  PFDNet in Sec~\ref{sec:persp_estim}.
At last, the learning objectives and the architecture of PFDNet are presented in Sec~\ref{sec:architecture}.

\subsection{Perspective-guided Fractional-dilation Convolution}\label{sec:PGFD}

Firstly, we give a brief introduction of perspective map.
Perspective maps encode the localized resolution of the scene in an image,
and the perspective value $\mathbf{s}^{gt}_j$ at each location $j$ represents the number of imaging pixels for a one-meter object at the location in the real scene~\cite{zhang2015cross}.
From~\cite{zhang2015cross, shi2019revisiting}, we have
\begin{equation}\label{eqn:s_gt}
\mathbf{s}^{gt} = \frac{h}{H}=\frac{1}{C-H}y_h,
\end{equation}
where $C$ is the camera height from the ground, $H$ is the height of a person on the ground, and the person's head top are mapped on the image plane at $y_h$.
$h$ is the observed person height and $H$ is set to the mean height of adults ($1.75m$).
Due to $C$ is fixed for each image, $\mathbf{s}^{gt}$ is a linear mapping of $y_h$ and remains the same in each row of $y_h$.
In~\cite{zhang2015cross}, the heights $h_i$ of several pedestrians at different positions has been manually labeled in each image, perspective value $p^{g}_j$ can be obtained by $p^{g}_j = \frac{h_j}{1.75}$ and $C$ can be estimated.
Finally, the entire ground-truth perspective map can be generated by employing a linear regression method to fit Eqn.~\eqref{eqn:s_gt}.

A straightforward way to deal with intra-scene variations is to adopt the larger receptive field for the people at larger scales, while to use the smaller one for that at smaller scales.
To this end, we propose a novel perspective-guided fractional-dilation convolution to adaptively adjust the receptive field of each convolution operation.
Perspective-guided fractional-dilation convolution has two appealing properties.
(1) By removing the limitation of discrete dilation rates, it extends regular dilation convolution to a more flexible form.
(2) It allocates spatially-variant receptive fields for different regions under the guidance of the dilation rate map generated from perspective maps.
In the following, we firstly introduce how to generate the dilation rate map from the perspective map, and then describe the definition and implementation of fractional-dilation convolution.
%

%
%
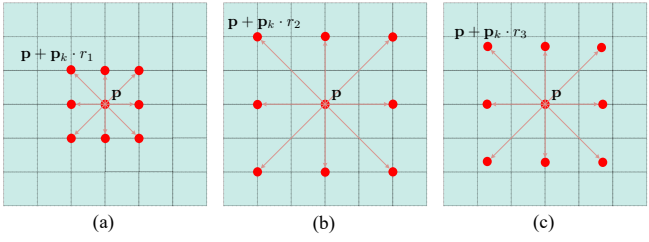
\begin{figure}[!t]
\centering
\begin{overpic}[scale=0.15]{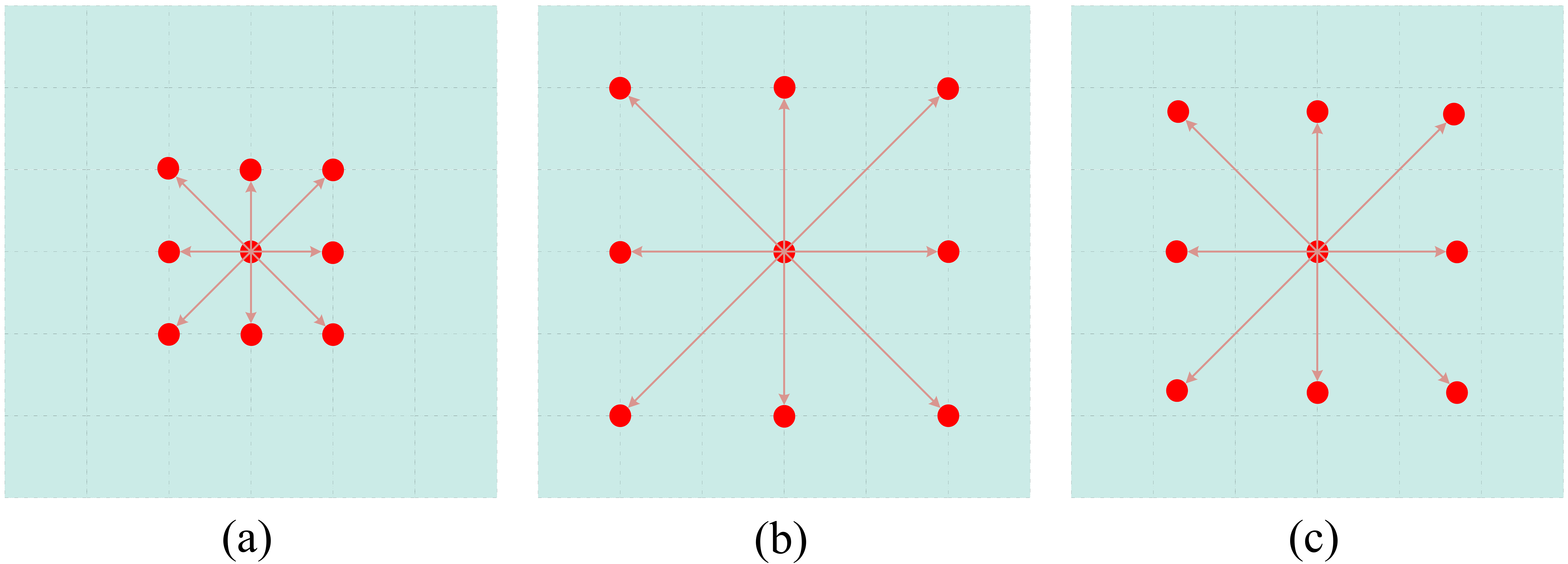}

\put(17, 21){\scalebox{.60}{\textcolor{black}{$\mathbf{p}$}}}
\put(51, 21){\scalebox{.60}{\textcolor{black}{$\mathbf{p}$}}}
\put(85, 21){\scalebox{.60}{\textcolor{black}{$\mathbf{p}$}}}

\put(3, 27){\scalebox{.60}{\textcolor{black}{$\mathbf{p} + \mathbf{p}_k\cdot r_1$}}}
\put(35, 32){\scalebox{.60}{\textcolor{black}{$\mathbf{p} + \mathbf{p}_k\cdot r_2$}}}
\put(70, 30.5){\scalebox{.60}{\textcolor{black}{$\mathbf{p} + \mathbf{p}_k\cdot r_3$}}}

\end{overpic}
 \scriptsize
\caption{Illustration of normal convolution, dilated convolution and factional-dilation convolution, as demonstrated in (a), (b) and (c). The red points indicates the sampling positions, the dashed lines show the grids and the intersections are grid points. Pixel/Feature values only exist on the grid points. When dilation rate $r$ is an integer (\eg, $r_1=1$ in (a) and $r_2=2$ in (b)), the sampling positions are on the grid points, thus feature aggregation can be directly performed. However, when $r$ is a fractional (\eg, $r_3$), we need to compute the values of sampling positions via bilinear interpolation.}
\label{fig:FDC}

\end{figure}

\subsubsection{Perspective-guided dilation rate map}\label{pg_dmap}
Let $\mathbf{x}$ and $\mathbf{s}$ be the feature maps and perspective map of an image, respectively.
Without loss of generality, $\mathbf{s}$ is downsampled to the same size $  H\times W$ with $\mathbf{x}$.
Following~\cite{shi2019revisiting}, $\mathbf{s}$ is firstly normalized to $[0,1]$ via a custom sigmoid-like function $\zeta(\cdot)$
\begin{equation}\label{eqn:ada_sig}
\widetilde{\mathbf{s}} = \zeta\left(\mathbf{s}\right)=\frac{1}{1+e^{-\alpha\left(\mathbf{s}-\beta\right)}}
\end{equation}
where $\alpha$ and $\beta$ are two trainable parameters.
Then the dilation rate map is defined as
\begin{equation}\label{eqn:dilation_rate_map}
    \mathbf{r}=\max(~\gamma * \widetilde{\mathbf{s}} + \theta, ~0~)
\end{equation}
where $\gamma$ and $\theta$ are another two parameters that should be learned in the training phase.
Here, we adopt $\mathbf{r}$ as the dilation rate map for fractional-dilation convolution.

\subsubsection{Fractional-dilation Convolution}
\label{sec:FDC}
A convolution filter with the kernel size $K \times K$ mainly accumulates the sampled values weighted by the filter parameters $\mathbf{w}$, where the sampled values are usually taken from a regular grid $\mathcal{R}$ on the input feature maps $\mathbf{x}$.
$\mathcal{R}$ enumerates all the integer points within the range $[- \lfloor K/2 \rfloor, \lfloor K/2 \rfloor]$.
%
%
In this way, for any position $\mathbf{p}$ in the output feature map $\mathbf{y}$, its feature representation can be defined as
\begin{equation}
\mathbf{y}(\mathbf{p}) = \sum_{\mathbf{p}_k \in \mathcal{R}}\mathbf{w}(\mathbf{p}_k)\cdot \mathbf{x}(\mathbf{p} + \mathbf{p}_k),
\end{equation}
where $\mathbf{p}$ and $\mathbf{p}_k$ indicate the original coordinate and offset coordinate, respectively.
Similarly, in a dilated convolution with dilation rate $r$, we have the following definition,
\begin{equation}
\label{eqn:navie_dilated_conv}
\mathbf{y}(\mathbf{p}) = \sum_{\mathbf{p}_k \in \mathcal{R}}\mathbf{w}(\mathbf{p}_k)\cdot \mathbf{x}(\mathbf{p} + \mathbf{p}_k \cdot r)
\end{equation}
where $r$ is a positive integer to make sure that the sampling positions are on the grid points.
Different from this normal definition, we propose fractional-dilation convolution of which the dilation rate can be a fractional number.
Fig.~\ref{fig:FDC} shows the differences between original convolution, dilated convolution and fractional-dilation convolution.

\begin{figure}[!htbp]
\centering
\begin{overpic}[scale=0.118]{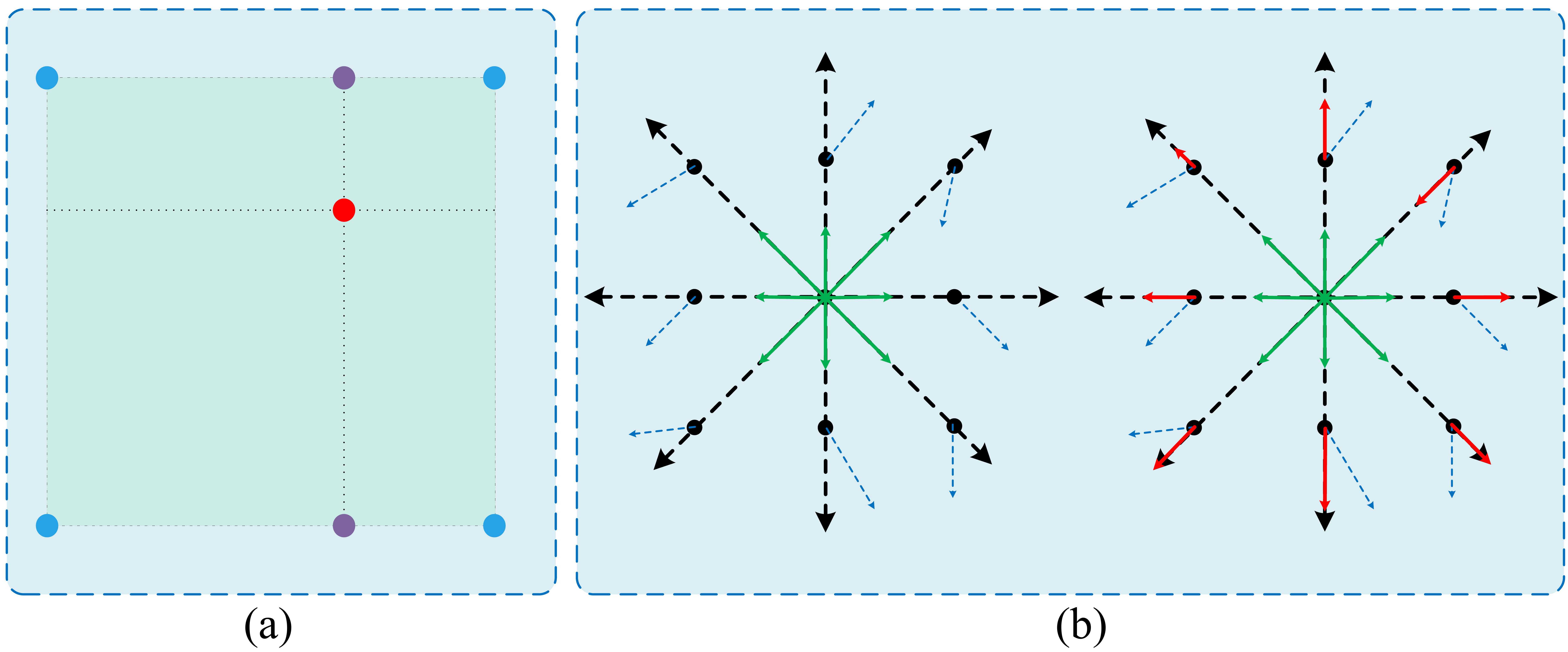}

\put(10, 34.5){\scalebox{0.5}{$\mu$}}
\put(4, 32){\scalebox{0.5}{$\lambda$}}

\put(24, 34.5){\scalebox{0.5}{$1 - \mu$}}
\put(4, 20){\scalebox{0.5}{$1 - \lambda$}}

\put(0.5, 38){\scalebox{0.5}{$(i, j)$}}
\put(26, 38){\scalebox{0.5}{$(i, j+1)$}}
\put(0.5, 5){\scalebox{0.5}{$(i+1, j)$}}
\put(23.5, 5){\scalebox{0.5}{$(i+1, j+1)$}}

\put(16, 38){\scalebox{0.5}{$(i, \hat{j})$}}
\put(14, 5){\scalebox{0.5}{$(i+1, \hat{j})$}}
\put(17, 26){\scalebox{0.5}{$(\hat{i}, \hat{j})$}}

\put(49.5, 28){\scalebox{0.7}{\textcolor{mygreen}{$\mathbf{p}_{k}$}}}

\put(81, 28){\scalebox{0.7}{\textcolor{mygreen}{$\mathbf{p}_{k}$}}}

\end{overpic}
 \scriptsize
\caption{Illustration of bilinear interpolation and gradient w.r.t. $r$ in fractional-dilation  convolution, where (a) shows the procedure of bilinear interpolation and subfigure (b) demonstrates the gradient with respect to $r$. In \textbf{subfigure (a)}, when $r$ is a fractional number, the sampling location $(\hat i, \hat j)$ is not on the grid point. We need to search the nearest four grid points and then perform linear interpolation in the horizontal direction to obtain $\mathbf{x}(i, \hat j)$ and $\mathbf{x}(i+1, \hat j)$. After that, we compute $\mathbf{x}(\hat i, \hat j)$ via linear interpolation in the vertical direction. In \textbf{subfigure (b)}, the blue dashed arrows indicate the derivatives with respect to sampling locations. The green arrows show the direction vectors, and the red arrows represent the projected vectors of blue vectors onto the corresponding green vectors.}
\label{fig:bilinear}

\end{figure}
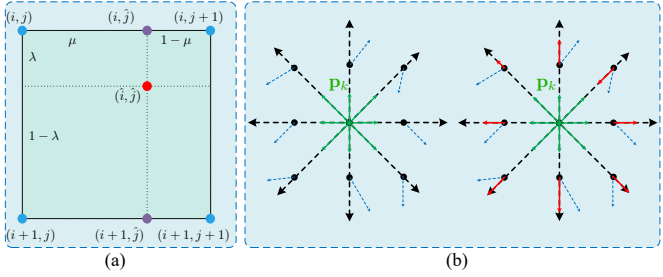

To realize Eqn.~\eqref{eqn:navie_dilated_conv} by using a fractional rate, we adopt bilinear interpolation to calculate the feature representation of the point with the fraction coordinate.
As illustrated in Fig.~\ref{fig:bilinear},  suppose the fraction coordinate of $\mathbf{\hat p}$ is $(\hat i ,\hat j)$, we firstly
search the nearest four grid points $(i,j)$, $(i,j\!+\!1)$, $(i\!+\!1,j)$  and $(i\!+\!1,j\!+\!1)$ respectively.
Then the bilinear interpolation is performed along the horizontal direction to calculate,
\begin{equation}
\begin{split}
\mathbf{x}(i, \hat j) &= \mu \cdot \mathbf{x}(i, j\!+\!1) + (1-\mu) \cdot \mathbf{x}(i, j) \\
\mathbf{x}(i\!+\!1, \hat j) &= \mu \cdot \mathbf{x}(i\!+\!1, j\!+\!1) + (1-\mu) \cdot \mathbf{x}(i\!+\!1, j) \\
\end{split}
\end{equation}
where $\mu$ is determined by the location of $(i,j)$ and $(i,j\!+\!1)$.
Then, we compute $\mathbf{x}(\hat i, \hat j)$ along vertical direction.
\begin{equation}
\label{eqn:detailed_bilinear}
\begin{split}
\mathbf{x}(\hat i, \hat j) &= \lambda \cdot \mathbf{x}(i\!+\!1, \hat j) + (1-\lambda) \cdot \mathbf{x}(i, \hat j) \\
&= \lambda \cdot \left[~ \mu \cdot \mathbf{x}(i\!+\!1, j\!+\!1) + (1-\mu) \cdot \mathbf{x}(i\!+\!1, j) ~\right] \\
&+ (1-\lambda) \cdot \left[~ \mu \cdot \mathbf{x}(i, j\!+\!1) + (1-\mu) \cdot \mathbf{x}(i, j)~ \right]
\end{split}
\end{equation}

From Fig.~\ref{fig:bilinear}(a), we have:
\begin{equation}
\lambda = 1-|(i+1)-\hat i|
\vspace{-0.6em}
\end{equation}
\begin{equation}
\mu = 1-|(j+1)-\hat j|
\end{equation}
And the coefficients of grid points in Eqn.~\eqref{eqn:detailed_bilinear} can be represented by:

\begin{equation}
\label{eqn:coeffi_1}
\lambda\mu = (1-|(i+1)-\hat i|)\cdot (1-|(j+1)- \hat j|)
\end{equation}
\begin{equation}
\label{eqn:coeffi_2}
\lambda(1-\mu) = (1-|(i+1)- \hat i|)\cdot (1-|j- \hat j|)
\end{equation}
\begin{equation}
\label{eqn:coeffi_3}
(1-\lambda)\mu = (1-|i- \hat i|)\cdot (1-|(j+1)-\hat j|)
\end{equation}
\begin{equation}
\label{eqn:coeffi_4}
(1-\lambda)(1-\mu) = (1-|i- \hat i|)\cdot (1-|j-\hat j|)
\end{equation}
%
%
Let $\Omega(\mathbf{\hat p})$ be the set of the nearest four grid points of $\mathbf{\hat p}$. $\mathbf{q}= (q_i,q_j)$ belongs to $\Omega(\mathbf{\hat p})$.
According to Eqn.~(\ref{eqn:detailed_bilinear}) and Eqn.~(\ref{eqn:coeffi_1}),~(\ref{eqn:coeffi_2}),~(\ref{eqn:coeffi_3}),~(\ref{eqn:coeffi_4}), we have the following simplified form,
\begin{equation}
\label{eqn:complex_bilinear}
\begin{split}
\mathbf{x}(\mathbf{\hat p}) = \sum_{\mathbf{q}\in \Omega(\mathbf{\hat p})} (1-|q_i-p_i|)\cdot ( 1-|q_j-p_j|)\cdot \mathbf{x}(\mathbf{q}).
\end{split}
\end{equation}
%
%
Then, we can further rewrite Eqn.~\eqref{eqn:complex_bilinear} as follows,
\begin{equation}\label{eqn:G}
    \mathbf{x}(\mathbf{\hat p}) = \sum_{\mathbf{q}\in \Omega(\mathbf{\hat p})}G(\mathbf{q}, \mathbf{\hat p}) \cdot \mathbf{x}(\mathbf{q}),
\end{equation}
where $\mathbf{\hat p}$ denotes an arbitrary location. According to Eqn.~(\ref{eqn:navie_dilated_conv}), we have $\mathbf{\hat p} = \mathbf{p} + \mathbf{p}_k \cdot r$.
And $\mathbf{q}$ enumerates integral spatial locations in the feature map $\mathbf{x}$, $G(\cdot, \cdot)$ denotes the bilinear interpolation kernel, which can be written as,
%
%
\begin{equation}
G(\mathbf{q}, \mathbf{\hat p}) = g(q_i, p_i) \cdot g(q_j, p_j),
\label{eqn:G}
\end{equation}
where $g(m, n) = \max (0, 1\!-\!|m \!-\! n|)$.
According to Eqn.~(\ref{eqn:G}), the output of such bilinear interpolation kernel  is non-zero for only a few neighbor points of $\mathbf{\hat p}$.
So we can obtain the meaningful feature representation of any fraction point.

%
%

\subsubsection{Back-propagation of Fractional-dilation Convolution}
In our fractional-dilation convolution, we optimize the filter weights and the dilation rate together.
The gradient with respect to the dilation rate $r$ is calculated by,
\begin{equation}
\begin{split}
& \frac{\partial{\mathbf{y}}(\mathbf{p})}{\partial{r}} = \sum_{\mathbf{p}_k \in \mathcal{R}} \mathbf{w}(\mathbf{p}_k)\cdot \frac{\partial{\mathbf{x}(\mathbf{p} + \mathbf{p}_k \cdot r)}}{\partial{r}} \\
&=  \sum_{\mathbf{p}_k \in \mathcal{R}}\left[
\mathbf{w}(\mathbf{p}_k) \cdot \sum_{\mathbf{q}\in \Omega(\mathbf{\hat p})} \frac{\partial{G}(\mathbf{q}, \mathbf{p} + \mathbf{p}_k \cdot r)}{\partial{r}} \cdot \mathbf{x}(\mathbf{q}) \right] \\
&= \sum_{\mathbf{p}_k \in \mathcal{R}}\left[
\mathbf{w}(\mathbf{p}_k) \cdot \mathbf{p}_k \cdot \sum_{\mathbf{q}\in \Omega(\mathbf{\hat p})} \frac{\partial{G}(\mathbf{q}, \mathbf{p} + \mathbf{p}_k \cdot r)}{\partial{(\mathbf{p} + \mathbf{p}_k \cdot r) }} \cdot \mathbf{x}(\mathbf{q})
\right]
\end{split}
\end{equation}
In fact, $\sum_{\mathbf{q}\in \Omega(\hat{\mathbf{p}})}
\frac{\partial{G}(\mathbf{q}, \mathbf{p} + \mathbf{p}_k \cdot r)}{\partial{(\mathbf{p} + \mathbf{p}_k \cdot r) }} \cdot \mathbf{x}(\mathbf{q})$ is 2D and denotes the derivative of bilinear interpolation with respect to the sampling location $\mathbf{p} + \mathbf{p}_k \cdot r$ (\ie, $\hat{\mathbf{p}}$).
We use the brief expression of \textit{the gradient with respect to sampling location} $\mathbf{p} + \mathbf{p}_k \cdot r$ to represent $\sum_{\mathbf{q}\in \Omega(\hat{\mathbf{p}})}\frac{\partial{G}(\mathbf{q}, \mathbf{p} + \mathbf{p}_k \cdot r)}{\partial{(\mathbf{p} + \mathbf{p}_k \cdot r) }} \cdot \mathbf{x}(\mathbf{q})$.
And $\mathbf{p}_k$ can represent a direction vector that starts from point $\mathbf{p}$ and ends at point $\mathbf{p} + \mathbf{p}_k$.
Intuitively, the formulation can be figuratively explained as Fig.~\ref{fig:bilinear}(b).
Here, $\mathbf{p}_k \cdot \sum_{\mathbf{q}\in \Omega(\hat{\mathbf{p}})} \frac{\partial{G}(\mathbf{q}, \mathbf{p} + \mathbf{p}_k \cdot r)}{\partial{(\mathbf{p} + \mathbf{p}_k \cdot r) }} \cdot \mathbf{x}(\mathbf{q})$
can be regarded as the result when the derivatives with respect to sampling locations are projected onto the corresponding direction vectors, shown as the red arrows in Fig.~\ref{fig:bilinear}.
If the result is positive/negative, the red arrow is in the same/opposite direction of the corresponding direction vector $\mathbf{p}_k$ and it encourages an expanding/shrinking of the dilation rate $r$.
%
%
Afterwards, the projected results are then weighted by the corresponding grid weight $\mathbf{w}(\mathbf{p}_k)$ and finally summed up as the final decision on scaling dilation rate $r$.
After obtaining the gradient of the dilation rate $r$ for each position, we could further calculate the parameters in  Eqn.~(\ref{eqn:ada_sig}) and Eqn.~(\ref{eqn:dilation_rate_map}) according to gradient chain rule.

%
%
\begin{figure}[t]

\centering

\begin{overpic}[scale=0.235]{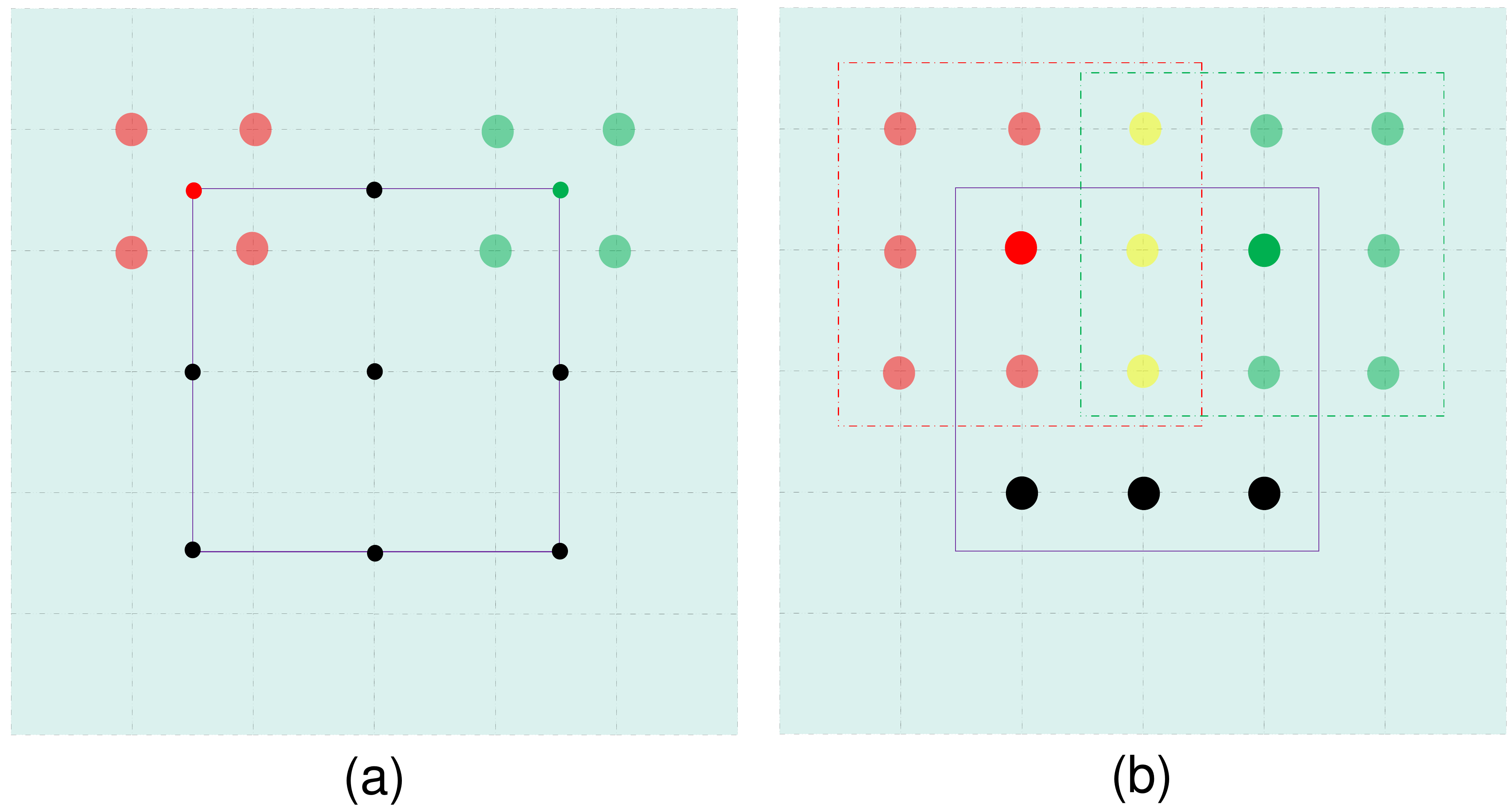}


\end{overpic}
 \scriptsize
\caption{Comparison on sampling positions of our proposed PFC and PGC~\cite{yan2019persp}. The dashed black lines show the grids and the intersections are grid points. Suppose we want to perform feature aggregation on a square region marked in purple. The region can be exactly covered by a fractional dilation convolution with dilation rate $r=1.5$.
The input feature map is termed as $\mathbf{x}$. (a) and (b) respectively show the procedure of PFC and PGC.
In (a), the sampling positions are marked with small balls and most of them are not on the grid points. Taken the upper-left red sampling position and the upper-right green sampling position for example, their values are estimated by the bilinear interpolation with the values (from $\mathbf{x}$) of two groups, four grid points per group. \emph{Obviously the two groups of points do not overlap with each other.} Finally, the feature aggregation is performed on the values of the small balls.
In (b), when conducting PGC on the region, we need to perform feature aggregation with the values (from the blurred feature $\mathbf{\widetilde{x}}$, instead of directly from input feature $\mathbf{x}$) of the grid points (\ie, here is $9$ big balls in the purple square). The value of red ball is obtained by performing Gaussian smoothing (with kernel size $3\times 3$ and $\sigma=\frac{3}{4}$) on the values (from $\mathbf{x}$) in the region of red square. Similarly, the value of green ball can be computed. The procedure is named as spatially variant Gaussian smoothing, and the blurred feature is represented as $\mathbf{\widetilde{x}}$. One can see that \emph{large overlap (colored in yellow) exists between the sampling points when performing Gaussian smoothing.} Furthermore, larger overlaps exist when the Gaussian kernel size goes bigger(\eg, it reaches $7\times 7$ in ~\cite{yan2019persp}), resulting in severe feature confusion.
}
\label{fig:pfc_vs_pgc}

\end{figure}

%
%
\begin{figure*}[!htbp]
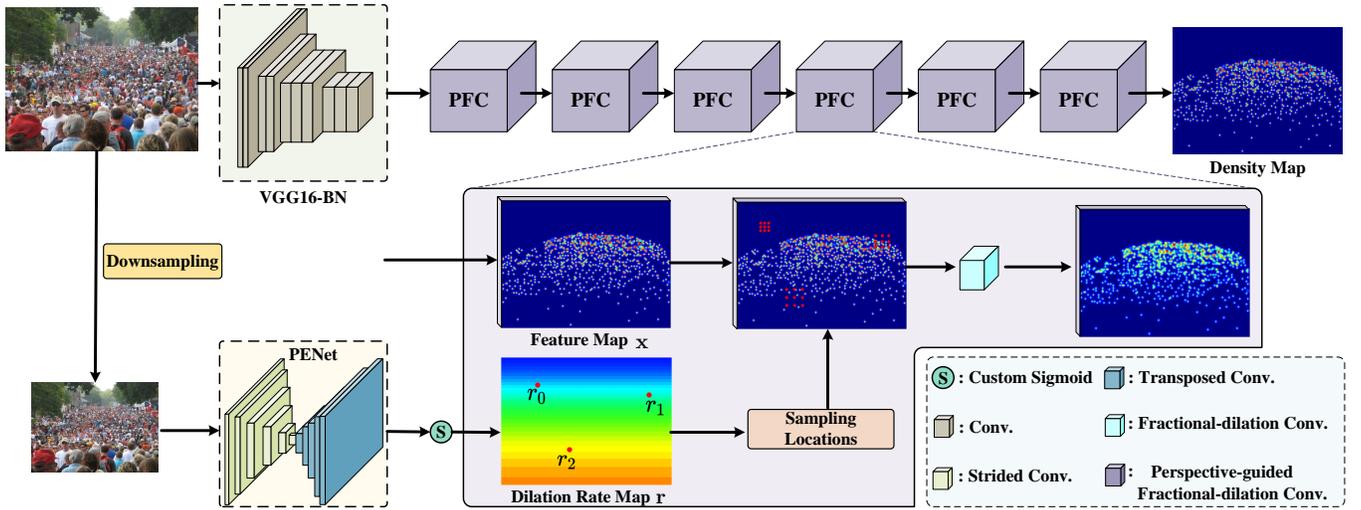

\centering
\begin{overpic}[scale=0.11]{./fig/net_arch/PFC_v7.pdf}
\put(48.5, 0.55){\scalebox{.80}{\textcolor{black}{$\mathbf{r}$}}}
\put(47.0, 12.1){\scalebox{.80}{\textcolor{black}{$\mathbf{x}$}}}

\end{overpic}
 \scriptsize
\caption{The architecture of the proposed PFDNet. PFNet aims to predict reasonable dilation rate map for each input image and then PFC takes the backbone features and estimated dilation rate map to perform spatially variant receptive field allocation. Six PFCs are embedded in our PFDNet.
The perspective map estimated by PENet is normalized via a custom sigmoid function to generate the dilation rate map. Afterwards, the sampled locations $\mathbf{p}+\mathbf{p}_k\cdot r$ for each point $\mathbf{p}$ on the feature map is obtained.}
\label{fig:PFDNet}

\end{figure*}

\subsubsection{Differences to PGC~\cite{yan2019persp}}
\label{sec:diff2pgc}
We firstly give a brief introduction to PGC.
PGC~\cite{yan2019persp} generates $\widetilde{s}$ via Eqn.~(\ref{eqn:ada_sig}), and then the blur map $\boldsymbol{\sigma}$ is obtained via linear mapping of $\widetilde{s}$, defined as
\begin{equation}\label{eqn:blur_map}
\boldsymbol{\sigma} = \max (a(\widetilde{s} - s_0), 0 ),
\end{equation}
where $a$ and $s_0$ are trainable parameters.
With spatially variant Gaussian smoothing, the smoothing result $\widetilde{x}_{i,j}$ is computed by
\begin{equation}\label{eqn:smoothing}
\widetilde{x}_{i,j} = \sum_k \sum_l x_{k,l} G_{\sigma_{i,j}} (i,j,k,l),
\end{equation}
where $G_{\sigma_{i,j}} (i,j,k,l)$ is defined as below:
\begin{equation}\label{eqn:gaussian}
\small
G_{\sigma_{i,j}} (i,j,k,l) \!=\! \frac{1}{\sqrt{2 \pi }\sigma_{i,j} } \exp \left( - \frac{\left( (k \!-\! i)^2 \!+\! (l \!-\! j)^2 \right)}{2\sigma_{i,j}^2}  \right).
\end{equation}
And the perspective-guided convolution is formulated as,
\begin{equation}\label{eqn:persp_conv}
\mathbf{y} = \mathbf{W}^{T} \widetilde{\mathbf{x}},
\end{equation}
where $\mathbf{W}$ denotes the conventional convolution kernel.
%

%
From Fig.~\ref{fig:pfc_vs_pgc}, it can be seen that PGC suffers from severe feature confusion due to large overlaps of sampling positions.
Intuitively, when the size of Gaussian kernel goes larger (\eg, in PGC~\cite{yan2019persp}, the maximum Gaussian kernel size is $7 \times 7$), larger overlap exists between the regions of Gaussian smoothing, which inevitably limits the performance of PGC.
%
%
By contrast, each sampling position of PFC is computed via bilinear interpolation of the nearest four grid points, and the groups of grid points hardly overlap with each other.
Therefore, PFC is effective in avoiding feature confusion.
Generally speaking, from the discussion in Sec.~\ref{sec:pgc}, our proposed PFC runs much faster, more flexible on allocating larger receptive field, can avoid feature confusion, and thus achieves better performance.

\subsection{Perspective Estimation}
\label{sec:persp_estim}
Although perspective maps are critical for calculating the dilation rate of fractional-dilation convolution, they are not always available in real applications.
To address such an issue, we introduce a perspective estimation branch to predict corresponding perspective information under the above situation.
Intuitively, one can pre-train a perspective estimation network on available pairs of image and perspective map, and then adopt it as the perspective estimation branch of the whole network.
However, we find a plain network optimized with the end-to-end manner usually achieves an unacceptable prediction accuracy.
%
%
Inspired by \cite{Mostajabi_2018_CVPR}, we propose a three-phase training strategy to train a robust auto-encoder for perspective estimation.

In the first phase, we use perspective maps from ShanghaiTech A~\cite{zhang2016single} to train an auto-encoder $D_{s}(E_{s}(\mathbf{s}; \Theta^E_s); \Theta^D_s)$ as a perspective re-constructor for each input $\mathbf{s}$.
Here, $\Theta^E_s$ and $\Theta^D_s$ denotes the parameters of the encoder $E_{s}$ and decoder $D_{s}$, respectively.
We adopt the $\ell_2$ reconstruction loss as the objective function,
\begin{equation}\label{eqn:persp_estimating_loss}
{\cal L}_{s2s} = \frac{1}{2N}\sum_{i=1}^{N}\| D_{s}(E_{s}(\mathbf{s}_i; \Theta^E_s);\Theta^D_s) - \mathbf{s}_i \|_{2}^{2} ,
\end{equation}
where $N$ is the total number of perspective maps.
Once the model is trained, the latent code $E_{s}(\mathbf{s}; \Theta^E_s)$ could encode the internal structure and contextual relationships in $\mathbf{s}$,
while the decoder $D_{s}$ could accurately recover a high quality perspective map from this latent code.

In the second phase, we train another auto-encoder which takes an image $\mathbf{I}$ as the input and predicts the corresponding perspective map $D_{s}(E_{I}(\mathbf{I}; \Theta^E_I); \Theta^D_s)$.
Specially, we reuse the decoder parameters $\Theta^D_s$ learned in the first phase, and only optimize the encoder parameters $\Theta^E_I$ by minimizing the following loss function,
\begin{equation}\label{eqn:persp_perceptual_loss}
{\cal L}_{I2s} = \frac{1}{2N}\sum_{i=1}^{N}\| D_{s}(E_{I}({I}_i; \Theta^E_I);\Theta^D_s) - \mathbf{s}_i \|_{2}^{2} .
\end{equation}

In the third phase, we transfer the optimized auto-encoder in the second phase onto the target training set.
The decoder is fixed during the training phase.
If the perspective maps are available on the target training set, we could optimize both ${\cal L}_{I2s}$ and density estimation loss (\ie, the main loss of the crowd counting task) together.
If the perspective maps are unavailable,  we could train the perspective estimation branch together with the main backbone network with an end-to-end manner.
%
%
In practice, benefited from the robustness of the decoder, even if the encoder is not well optimized, the decoder can still recover a roughly reasonable perspective map.
According to the above three training phases, our perspective estimation branch can be optimized in a weakly-supervised way even without using the corresponding perspective annotations.

\subsection{Network Architecture}
\label{sec:architecture}
Fig.~\ref{fig:PFDNet} illustrates the architecture of our PFDNet, which is comprised of the main backbone network together with the perspective estimation network (PENet).
We adopt CSRNet~\cite{li2018csrnet} as our main backbone network, and replace all of the regular dilated convolutions with our proposed PFCs.
The dilation rate map for the PFC are calculated based on the output of PENet according to the Eqn.~(\ref{eqn:ada_sig}) and Eqn.~(\ref{eqn:dilation_rate_map}).
The detailed PENet architecture is illustrated in Table~\ref{table:PENet}.

\begin{table}[!ht]
   \begin{center}
      \scriptsize
	 \caption{The architecture of PENet. ``LReLU'' denotes leaky ReLU with the slope of 0.2. ``UpConv'' represents transposed convolution. $H$ and $W$ are the height and width of input respectively.}
\label{table:PENet}
    \resizebox{0.95\hsize}{!}{
      \begin{tabular}{ c| c |c }
    \hline
      Part & Layer & Output Size \\
      \midrule
      \multirow{6}*{Encoder}
     & Conv. (3, 3, 32), stride=2; LReLU & $32\times H/2\times W/2$ \\
     & Conv. (3, 32, 64), stride=2; LReLU & $64\times H/4\times W/4$ \\
     & Conv. (3, 64, 128), stride=2;  LReLU & $128\times H/8\times W/8$ \\
     & Conv. (3, 128, 256), stride=2;  LReLU & $256\times H/16\times W/16$ \\
     & Conv. (3, 256, 512), stride=2;  LReLU & $512\times H/32\times W/32$ \\
     & Conv. (3, 512, 1024), stride=2;  LReLU & $1024\times H/64\times W/64$ \\
      \hline\hline
     & UpConv. (3, 1024, 512), stride=2; ReLU & $512\times H/32\times W/32$ \\
      \multirow{6}*{Decoder}
     & UpConv. (3, 512, 256), stride=2; ReLU & $256\times H/16\times W/16$ \\
     & UpConv. (3, 256, 128), stride=2; ReLU & $128\times H/8\times W/8$ \\
     & UpConv. (3, 128, 64), stride=2; ReLU & $64\times H/4\times W/4$ \\
     & UpConv. (3, 64, 32), stride=2; ReLU & $32\times H/2\times W/2$ \\
     & UpConv. (3, 3, 1), stride=2;  ReLU & $1\times H\times W $ \\
      \bottomrule
  	\end{tabular}
  	}
  	\end{center}

\end{table}

%

%
\section{Experimental Results and Dataset}
\label{sec:experiments_results}
In this section, we will first give the description of the implementation details, and then present the comparison between state-of-the-arts and our PFDNet on five datasets, namely ShanghaiTech~\cite{zhang2016single}, WorldExpo'10~\cite{zhang2015cross}, UCF-QNRF~\cite{idrees2018composition}, UCF\_CC\_50~\cite{idrees2013multi} and TRANCOS~\cite{guerrero2015extremely}. Extensive ablation study is then conducted to clarify the contribution  of each component in PFDNet.
The pre-trained models and source code are online available at~\url{https://github.com/Zhaoyi-Yan/PFDNet}.

\subsection{Evaluation Metric and Dataset}
We adopt Mean Absolute Error (MAE) and Root Mean Square Error (RMSE) as the evaluation metrics, the same as previous works~\cite{zhang2016single, sam2017switching, li2018csrnet}.
Let $N$ be the number of test samples, $C_i$ and $\hat{C}_i$ are respectively the ground-truth count and prediction one.
Then MAE and RMSE can be defined as follows:
\begin{equation}\label{eqn:mae}
MAE = \frac{1}{N}\sum_{i=1}^{N}|C_i - \hat{C}_{i}|,
\end{equation}

\begin{equation}\label{eqn:mse}
RMSE = \sqrt{\frac{1}{N}\sum_{i=1}^{N}\|C_i - \hat{C}_{i}\|^2}.
\end{equation}

%
We evaluate the performance of the proposed PFDNet and other state-of-the-art methods on five popular benchmarks. The detailed descriptions are as follows.

\textbf{ShanghaiTech} contains $1,198$ images with a total of $330,165$ annotated people.
ShanghaiTech A contains $482$ crowd images with crowd numbers varying from $33$
to $3139$, and ShanghaiTech B contains $716$ high-resolution
images with crowd numbers from $9$ to $578$.
For Part A, the images are collected from the web, while the images in Part B are Shanghai street views.

\textbf{WorldExpo'10} contains $3,980$ images from the $2010$ Shanghai WorldExpo.
The training set contains $3,380$ images, while the test set is divided into five different scenes with $120$ images each.
ROIs are provided to indicate the target regions for training / testing.
Following~\cite{li2018csrnet}, each image and its ground-truth density map are masked with the ROI in preprocessing.

\textbf{UCF-QNRF} is a large crowd counting dataset with $1,535$ high-resolution images and $1.25$ million head annotations.
There are $1,201$ training images and the rest are used as test images.
It contains extremely congested scenes where the maximum count of an image $12,865$.

\textbf{UCF\_CC\_50} contains $50$ images of diverse scenes, which is a challenging dataset.
The head count per image varies drastically (from $94$ to $4,543$).
Following~\cite{idrees2013multi}, we split the dataset into five subsets and perform a $5$-fold cross-validation.

\textbf{TRANCOS} is a public traffic dataset containing $1,244$ images of different congested traffic scenes captured
by surveillance cameras with $46,796$ annotated vehicles.
The regions of interest (ROI) are provided for evaluation.
We evaluate PFDNet on TRANCOS dataset to validate the generalization of our method and the robustness of
PENet.
The Grid Average Absolute Error (GAME) is adopted as the evaluation metric, which can be referred in~\cite{guerrero2015extremely, li2018csrnet}.
GAME(L) subdivides the image using a grid of $4^L$ non-overlapping regions, and the error is computed as the sum of the mean absolute errors in each of these regions.
Obviously, GAME(0) is equivalent to the MAE metric.

\subsection{Implementation Details}
\label{sec:training_details}
Let $\Theta$ denote the model parameters of PFDNet $\Phi(I; \Theta)$. Then the main loss function of the crowd counting estimation task is  as follows,
\begin{equation}\label{eqn:pmae_pmse}
{\cal L}\left(\Theta\right) = \frac{1}{2N}\sum_{i=1}^{N}\|\Phi(I; \Theta) - Y_{i}\|_{2}^{2},
\end{equation}
where $Y_i$ is the ground-truth density map of $I_i$.
We generate ground-truth density maps by adopting a normalized Gaussian kernel to blur each head annotation, and make sure the summations equals to the crowd counts.
In our experiments, we use a fixed $15\!\times\!15$ Gaussian kernel with standard deviation of $4$ to generate density maps.
%

We adopt Adam~\cite{kingma2014adam} as the optimizer.
The learning rate is set as $1e^{-4}$ for the parameters of backbone network.
Original CSRNet perfers one whole image in an iteration when training, consuming a lot of time for some datasets (\eg, ShanghaiTech A which contains images of various sizes).
To make the training procedure more efficient, we adopt VGG16-BN~\cite{simonyan2014very} model instead of VGG-16 as the backbone, and add extra BatchNorm~\cite{ioffe2015batch} layers after each dilated convolution in CSRNet.
BatchNorm layer stabilizes the model when training with random cropping images and gets rid of performance degradation when evaluating.
Besides, we upsample the predicted density map by the factor 4, making the final predicted density map $\frac{1}{2}$ of the original input size.
\textbf{After all the changes, we call it CSRNet* and take it as the baseline of our method}.
It should be noted that CSRNet* shows its superiority in faster training (\ie, batch size can be larger than 1, and is 16 in our experiments.) and higher performances than original CSRNet on the validation datasets.
The weights of PFDNet are initialized with Gaussian distribution of zero mean and 0.01 standard deviation, except the weights of backbone are initialized with the VGG16-BN pretrained on ImageNet.
We implement our PFDNet on Pytorch~\cite{paszke2017automatic}.
Random flipping and color jittering are adopted for data augmentation.

For the initialization of $\alpha$, $\beta$, $\gamma$ and $\theta$, according to CSRNet~\cite{li2018csrnet}, the average dilated rate over the whole map is expected to be around $2$.
Empirically, we set $\alpha$, $\beta$, $\gamma$ and $\theta$ respectively with value $1$, $1$, $1.5$, $1$ for all the PFC layers.
For simplicity, we adopt the same initialization values for these four super-parameters on all the datasets.

For ShanghaiTech and WorldExpo'10, since the corresponding ground-truth perspective maps are available, they are used as the guidance of PFC directly, and our model is trained for $300$ epochs without PENet.
%
For the datasets without perspective annotations, the PENet is pre-trained for the first two phases of Sec.~\ref{sec:persp_estim}, both using $500$ epochs. After that the main network and PENet are jointly trained for another $300$ epochs, following the description of the third phase of Sec.~\ref{sec:persp_estim}.
In practice, we build PFDNet by stacking six PFC layers after a truncated VGG16-BN for all datasets.
It takes about $2$ days to train the network on UCF-QNRF with an NVIDIA Tesla P100 GPU.

\subsection{Evaluations and Results}
\label{sec:eval_and_comparison}
Five datasets are adopted in our experiments, including ShanghaiTech~\cite{zhang2016single}, WorldExpo'10~\cite{zhang2015cross}, UCF-QNRF~\cite{idrees2018composition}, UCF\_CC\_50~\cite{idrees2013multi} and TRANCOS~\cite{guerrero2015extremely}.
It is noted that the first two contain available perspective maps\footnote{Annotations of ShanghaiTech are from ~\cite{shi2019revisiting} and WorldExpo'10 provides official annotations.}, thus we can directly use these annotations to train our PFDNet.
For the later three datasets that perspective map annotations are unavailable, we denote \textbf{\emph{Ours A}} as directly adopting the estimated perspective maps (based on the PENet trained on perspective annotations from ShanghaiTech A in the first two phases) as \emph{ground-truth} and feed them to the training of PFC layers;
while \textbf{\emph{Ours B}} as the end-to-end training without perspective map annotations in the third training phase described in Sec.~\ref{sec:persp_estim}, where the main network and PENet are jointly trained without any perspective map annotations.
As the resolution of perspective map required by PFCs is only $1/8$ size of the original image,
thus all input images are firstly downsampled to $1/8$ resolution of the original size in the two training phases of PENet.

\begin{figure}[!t]
\centering
\includegraphics[width=0.5\textwidth]{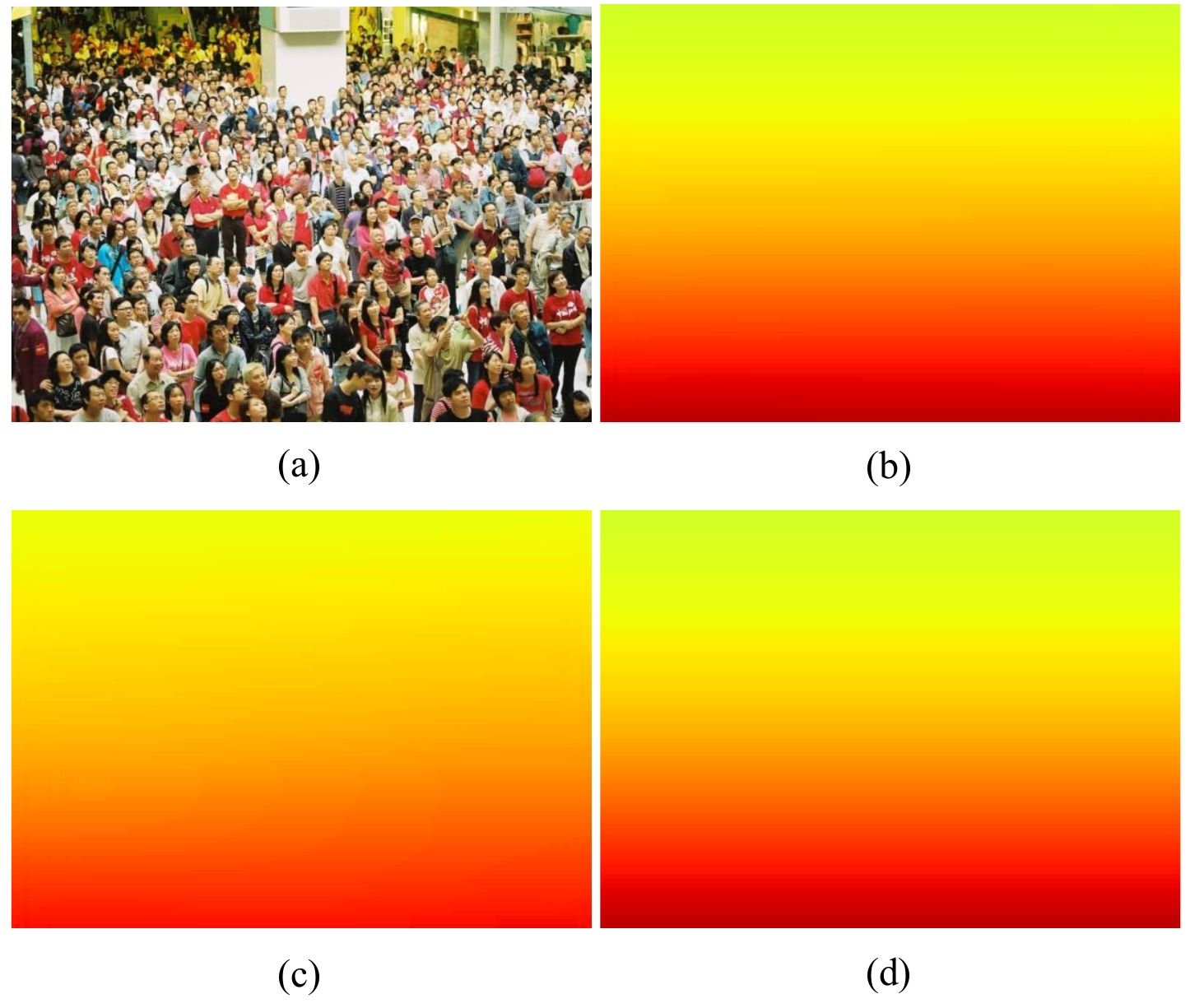}
 \scriptsize
\caption{Perspective estimations of PENet in the first two phases. (a) is the input; (b)(c) are respectively the predictions in the first and second phase; (d) is the ground-truth perspective annotations.}
\label{fig:PENet_out}

\end{figure}

For \emph{Ours A} / \emph{Ours B}, by following Sec.~\ref{sec:persp_estim} with ShanghaiTech A, we get $30.6$ dB PSNR in the first phase and $25.1$ dB PSNR in the second phase.
Besides, Fig.~\ref{fig:PENet_out} shows a test sample for visualization of the perspective estimation.
Fig.~\ref{fig:PENet_out}(b)(d) shows the quite similar estimation results, which indicates the auto-encoder is well trained in the first phase.
In the second phase, the auto-encoder aims at predicting perspective estimations (\eg, Fig.~\ref{fig:PENet_out}(c)) from RGB images.
It is observed that Fig.~\ref{fig:PENet_out}(c) is still visually-satisfying, indicating that the PENet could actually recover good perspective annotations from a RGB image.

%
%

\begin{table*}[!t]
\footnotesize
\centering
\small
\caption{Comparisons on ShanghaiTech A, ShanghaiTech B and WorldExpo'10. It is noted that we adopt \textbf{ground-truth} perspective annotations to feed PFCs. The results of top two performance are highlighted in red and blue respectively.}
\resizebox{0.95\hsize}{!}{
 	\begin{tabular}{
 	l|*1{p{0.8cm}<{\centering}}|*1{p{1cm}<{\centering}}|*2{p{0.8cm}<{\centering}}|*2{p{0.8cm}<{\centering}}|*6{p{0.8cm}<{\centering}}
 	}
		\toprule
		{\multirow{2}{*}{\textbf{Method}}} & {\multirow{2}{*}{\textbf{Year}}} & {\multirow{2}{*}{\textbf{Venue}}} &
		\multicolumn{2}{c|}{\textbf{ShanghaiTech A}} & \multicolumn{2}{c|}{\textbf{ShanghaiTech B}} & \multicolumn{6}{c}{\textbf{WorldExpo'10}}  \\
 		\cline{4-13}
		{} & {} & {} & MAE & RMSE & MAE & RMSE & Sce.1 & Sce.2 & Sce.3 & Sce.4 & Sce.5 & Avg \\
		\hline
		\hline
		CSRNet~\cite{li2018csrnet} & 2018 & CVPR & 68.2 & 115.0 & 10.6 & 16.0 & 2.9 & 11.5 & 8.6 & 16.6 & 3.4 & 8.6 \\
		SANet~\cite{cao2018scale} & 2018 & CVPR & 67.0 & 104.5 & 8.4 & 13.6 & 2.6 & 13.2 & 9.0 & 13.3 & 3.0 & 8.2 \\
		HA-CCN~\cite{sindagi2019ha} & 2019 & TIP & 62.9 & 94.9 & 8.1 & 13.4 & - & - & - & - & - & - \\
		TEDnet~\cite{jiang2019crowd} & 2019 & CVPR & 64.2 & 109.1 & 8.2 & 12.8 & 2.3 & 10.1 & 11.3 & 13.8 & 2.6 & 8.0 \\
		ADCrowdNet~\cite{liu2019adcrowdnet} & 2019 & CVPR & 63.2 & 98.9 & 8.2 & 15.7 & 1.6 & 13.2 & 8.7 & 10.6 & 2.6 & 7.3 \\
		SFCN~\cite{wang2019learning} & 2019 & CVPR & 65.2 & 109.4 & 7.2 & 12.2 & - & - & - & - & - & - \\
		PACNN~\cite{shi2019revisiting} & 2019 & CVPR & 62.4 & 102.0 & 7.6 & 11.8 & 2.3 & 12.5 & 9.1 & 11.2 & 3.8 & 7.8 \\
		CAN~\cite{liu2019context} & 2019 & CVPR & 62.3 & 100.0 & 7.2 & 11.1 & 2.9 & 12.0 & 10.0 & 7.9 & 4.3 & 7.4 \\
		PSDDN~\cite{liu2019point} & 2019 & CVPR & 65.9 & 112.3 & 9.1 & 14.2 & - & - & - & - & - & - \\
		SPN~\cite{Xu2019learn} & 2019 & ICCV & 64.2 & 98.4 & 7.2 & 11.1 & - & - & - & - & - & - \\
		BL~\cite{liu2019context} & 2019 & ICCV & 62.3 & 100.0 & 7.2 & 11.1 & - & - & - & - & - & - \\
		S-DCNet~\cite{xhp2019SDCNet} & 2019 & ICCV & 58.3 & 95.0 & \textcolor{blue}{6.7} & \textcolor{blue}{10.7} & - & - & - & - & - & - \\
		PGCNet~\cite{yan2019persp} & 2019 & ICCV & 57.0 & 86.0 & 8.8 & 13.7 & 2.5 & 12.7 & 8.4 & 13.7 & 3.2 & 8.1 \\
		DensityCNN~\cite{jiang2020density} & 2020 & TMM & 63.1 & 106.3 & 9.1 & 16.3 & 1.8 & 11.2 & 10.1 & 7.9 & 3.4 & \textcolor{blue}{6.9} \\
		DENet~\cite{liu2020denet} & 2020 & TMM & 65.5 & 101.2 & 9.6 & 15.4 & 2.8 & 10.7 & 8.6 & 15.2 & 3.5 & 8.2 \\
		EPA~\cite{yang2020embedding} & 2020 & TIP & 60.9 & 91.6 & 7.9 & 11.6 & 2.1 & 9.3 & 9.4 & 11.0 & 3.7 & 7.4 \\	
		KDMG~\cite{wan2020kernel} & 2020 & TPAMI & 63.8 & 99.2 & 7.8 & 12.7 & - & - & - & - & - & - \\	
		LSC-CNN~\cite{sam2020locate} & 2020 & TPAMI & 66.4 & 117.0 & 8.1 & 12.7 & 2.9 & 11.3 & 9.4 & 12.3 & 4.3 & 8.0 \\
		LCM~\cite{hu2020count} & 2020 & ECCV & 61.6 & 98.4 & 7.0 & 11.0 & - & -  & - & - & - & -  \\
		AMSNet~\cite{liu2020adaptive} & 2020 & ECCV & \textcolor{blue}{56.7} & \textcolor{blue}{93.4} & \textcolor{blue}{6.7} & \textcolor{red}{10.2} & 1.6 & 8.8 & 10.8 & 10.4 & 2.5 & \textcolor{red}{6.8}  \\
		DM-Count~\cite{wang2020DMCount} & 2020 & NeurIPS & 61.9 & 99.6 & 7.4 & 11.8 & - & - & - & - & - & -  \\
		\hline
		\hline
		Baseline & - & - & 65.1 & 112.4 & 9.8 & 15.2 & 2.7 & 11.1 & 8.5 & 16.7 & 3.6 & 8.5 \\
		Ours & - & - & \textcolor{red}{\textbf{53.8}} & \textcolor{red}{\textbf{89.2}} & \textcolor{red}{\textbf{6.5}} & \textcolor{blue}{\textbf{10.7}} & \textbf{2.5} & \textbf{9.7} & \textbf{8.1} & \textbf{10.1} & \textbf{3.5} & \textcolor{red}{\textbf{6.8}} \\
		\bottomrule
	\end{tabular}
}
\vspace{1em}
\label{table:sha_shb_we}
\end{table*}

\begin{table*}[!t]
\footnotesize
\centering
\small
\caption{Comparisons on UCF-QNRF, UCF\_CC\_50 and TRANCOS. Baseline is the model CSRNet*, and \textit{Ours A} means we directly use the estimated perspective maps predicted by PENet. While \textit{Ours B} indicates that PENet embeded as a branch and the whole training is end-to-end. The results of top two performance are highlighted in red and blue respectively.}
\resizebox{0.95\hsize}{!}{
 	\begin{tabular}{
 	l|*1{p{0.8cm}<{\centering}}|*1{p{1cm}<{\centering}}|*2{p{0.8cm}<{\centering}}|*2{p{0.8cm}<{\centering}}|*4{p{0.8cm}<{\centering}}
 	}
		\toprule
		{\multirow{2}{*}{\textbf{Method}}} & {\multirow{2}{*}{\textbf{Year}}} & {\multirow{2}{*}{\textbf{Venue}}} &
		\multicolumn{2}{c|}{\textbf{UCF-QNRF}} & \multicolumn{2}{c|}{\textbf{UCF\_CC\_50}} & \multicolumn{4}{c}{\textbf{TRANCOS}}  \\
 		\cline{4-11}
		{} & {} & {} & MAE & RMSE & MAE & RMSE & GE(0) & GE(1) & GE(2) & GE(3) \\
		\hline
		\hline
		CSRNet~\cite{li2018csrnet} & 2018 & CVPR & - & - & 266.1 & 397.5 & 3.56 & 5.49 & 8.57 & 15.04 \\
		SANet~\cite{cao2018scale} & 2018 & CVPR & - & - & 258.4 & 334.9 & - & - & - & - \\
		HA-CCN~\cite{sindagi2019ha} & 2019 & TIP & 118.1 & 180.4 & 256.2 & 348.4 & - & - & - & - \\
		TEDnet~\cite{jiang2019crowd} & 2019 & CVPR & 113 & 188 & 294.4 & 354.5 & - & - & - & -  \\
		ADCrowdNet~\cite{liu2019adcrowdnet} & 2019 & CVPR & - & - & 257.1 & 363.5 & - & - & - & -  \\
		SFCN~\cite{wang2019learning} & 2019 & CVPR & 102.0 & 171.4 & 214.2 & 318.2 & - & - & - & -  \\
		PACNN~\cite{shi2019revisiting} & 2019 & CVPR & - & - & 241.7 & 320.7 & - & - & - & -  \\
		CAN~\cite{liu2019context} & 2019 & CVPR & 107 & 183 & 212.2 & \textcolor{red}{243.7} & - & - & - & -  \\
        PSDDN~\cite{liu2019point} & 2019 & CVPR & - & - & 359.4 & 514.8 & 4.79 & 5.43 & 6.68 & 8.40 \\
		SPN~\cite{Xu2019learn} & 2019 & ICCV & 104.7 & 173.6 & 259.2 & 335.9 & - & - & - & -  \\
		BL~\cite{liu2019context} & 2019 & ICCV & 88.7 & 154.8 & 229.3 & 308.2 & - & - & - & -  \\
		S-DCNet~\cite{xhp2019SDCNet} & 2019 & ICCV & 104.4 & 176.1 & 204.2 & 301.3 & - & - & - & -  \\
		PGCNet~\cite{yan2019persp} & 2019 & ICCV & - & - & 244.6 & 361.2 & - & - & - & -  \\
		DensityCNN~\cite{jiang2020density} & 2020 & TMM & 101.5 & 186.9 & 244.6 & 341.8 & \textcolor{blue}{3.17} & 4.78 & \textcolor{blue}{6.30} & \textcolor{red}{8.26}  \\
		DENet~\cite{liu2020denet} & 2020 & TMM & - & - & 241.9 & 345.4 & - & - & - & -  \\
		LSC-CNN~\cite{sam2020locate} & 2020 & TPAMI & 120.5 & 218.2 & 225.6 & 302.7 & 4.6 & 5.4 & 6.9 & \textcolor{blue}{8.3}  \\
		EPA~\cite{yang2020embedding} & 2020 & TIP & - & - & 250.1 & 342.1 & 3.21 & - & - &  - \\
		KDMG~\cite{wan2020kernel} & 2020 & TPAMI & 99.5 & 173.0 & - & - & 3.13 & 4.79 & 6.20 &  8.68 \\
		LCM~\cite{hu2020count} & 2020 & ECCV & 86.6 &  152.2 & \textcolor{red}{184.0} & \textcolor{blue}{265.8} & - & -  & - & -   \\
		AMSNet~\cite{liu2020adaptive} & 2020 & ECCV & 101.8 & 163.2 & 208.4 & 297.3 & - & - & - & -   \\
		DM-Count~\cite{wang2020DMCount} & 2020 & NeurIPS & \textcolor{blue}{85.6} & 148.3 & 211.0 & 291.5 & - & - & - & -   \\
		\hline
		\hline
		Baseline & - & - & 103.1 & 188.6 & 257.1 & 363.8 & 3.49 & 5.30 & 8.16 & 14.45 \\
		Ours A & - & - & 87.5  & 152.4 & 224.6 & 304.4 & 3.22 & \textcolor{blue}{4.58} & 6.74 & 10.69 \\
		Ours B & - & - & \textcolor{red}{\textbf{84.3}} & \textcolor{red}{\textbf{141.2}} & \textcolor{blue}{\textbf{205.8}} & \textbf{289.3} & \textcolor{red}{\textbf{3.06}} & \textcolor{red}{\textbf{4.15}} & \textcolor{red}{\textbf{6.22}} & \textbf{9.86} \\
		\bottomrule
	\end{tabular}
}
\vspace{1em}
\label{table:qnrf_ucf_trans}
\end{table*}

\begin{figure*}[!t]
\centering
\includegraphics[width=1.0\textwidth]{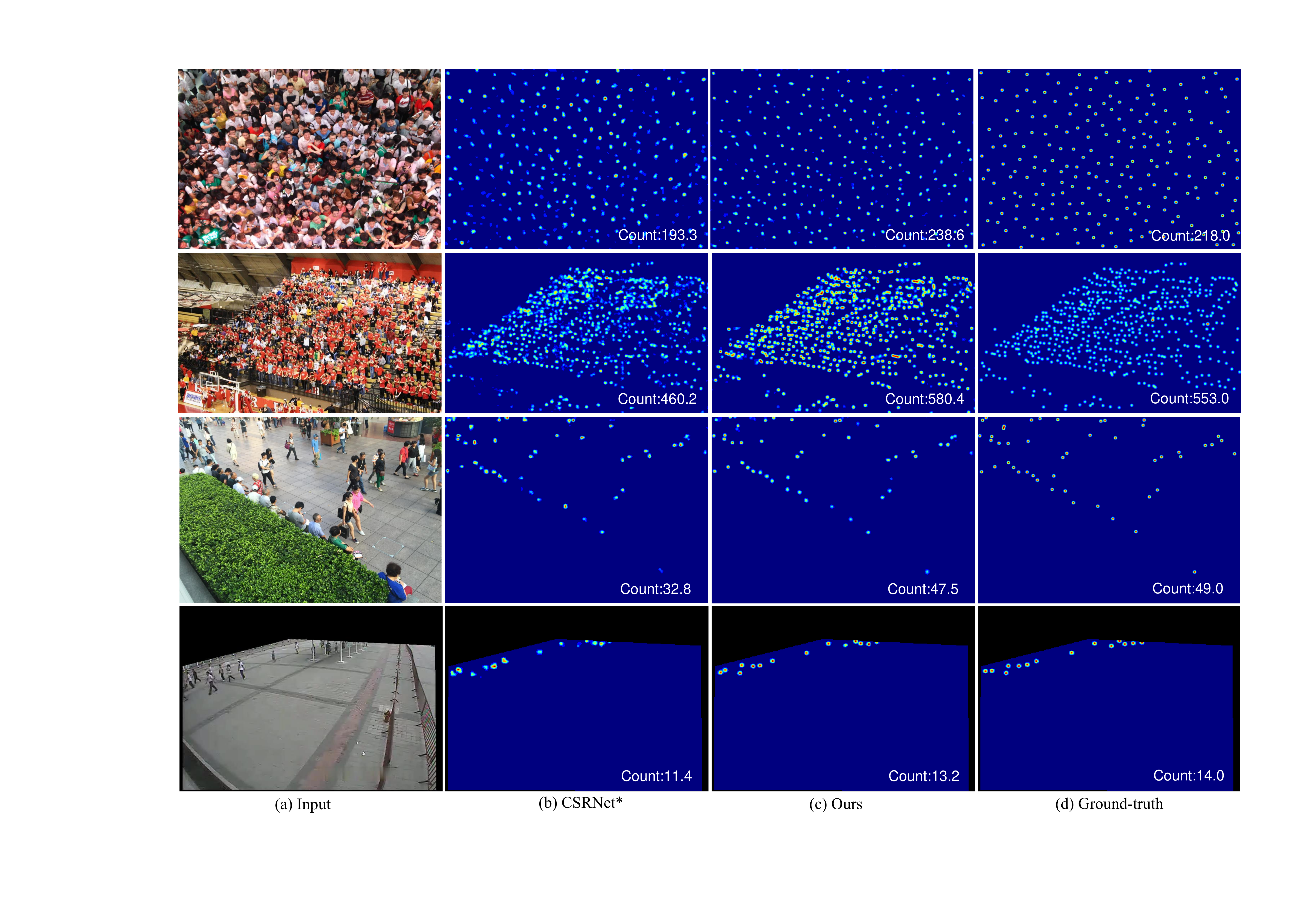}
 \scriptsize
\caption{Density maps estimated by CSRNet* and Ours. It is seen that our PFDNet delivers superior performance over CSRNet* in both estimating better density maps and more accurate counts.}
\label{fig:sample}

\end{figure*}

Table~\ref{table:sha_shb_we} 
shows the results on ShanghaiTech and WorldExpo'10, in which we feed ground-truth perspective annotations to the PFCs.
It is seen that our method shows significant performance gain against the baseline CSRNet* with $17.4\%$ relative MAE decrease on ShanghaiTech A and $33.7\%$ relative MAE decrease on ShanghaiTech B.
For WorldExpo'10 dataset, our method achieves the best $6.8$ average MAE with $20.0\%$ performance improvement.
For all these three datasets, our method performs the best, indicating the superiority of spatially variant receptive field allocation.
Besides, some test cases can be found in Fig.~\ref{fig:sample}, clearly indicating that the superiority of PFDNet to CSRNet* in estimating a better density map.

For the datasets UCF-QNRF, UCF\_CC\_50 and TRANCOS, they do not provide official perspective annotations.
Thus, we conduct the two comparison experiments \emph{Ours A} and \emph{Ours B} described at the beginning of this section.
Table~\ref{table:qnrf_ucf_trans} shows the results of these three datasets.
Compared with the baseline CSRNet*, \emph{Ours A} achieves significant gain on both MAE and RMSE.
Specifically, $15.7\%$, $18.1\%$ performance gains on MAEs for datasets UCF-QNRF and UCF\_CC\_50, achieving $87.5$ and $224.5$ MAE.
Notably, \emph{Ours B} further reduces the MAEs, ranks first on UCF-QNRF and achieves the second best performance on UCF\_CC\_50, which shows the feasibility of our end-to-end training strategy described in Sec.~\ref{sec:persp_estim}.

\begin{figure}[!t]
\centering
\includegraphics[width=0.5\textwidth]{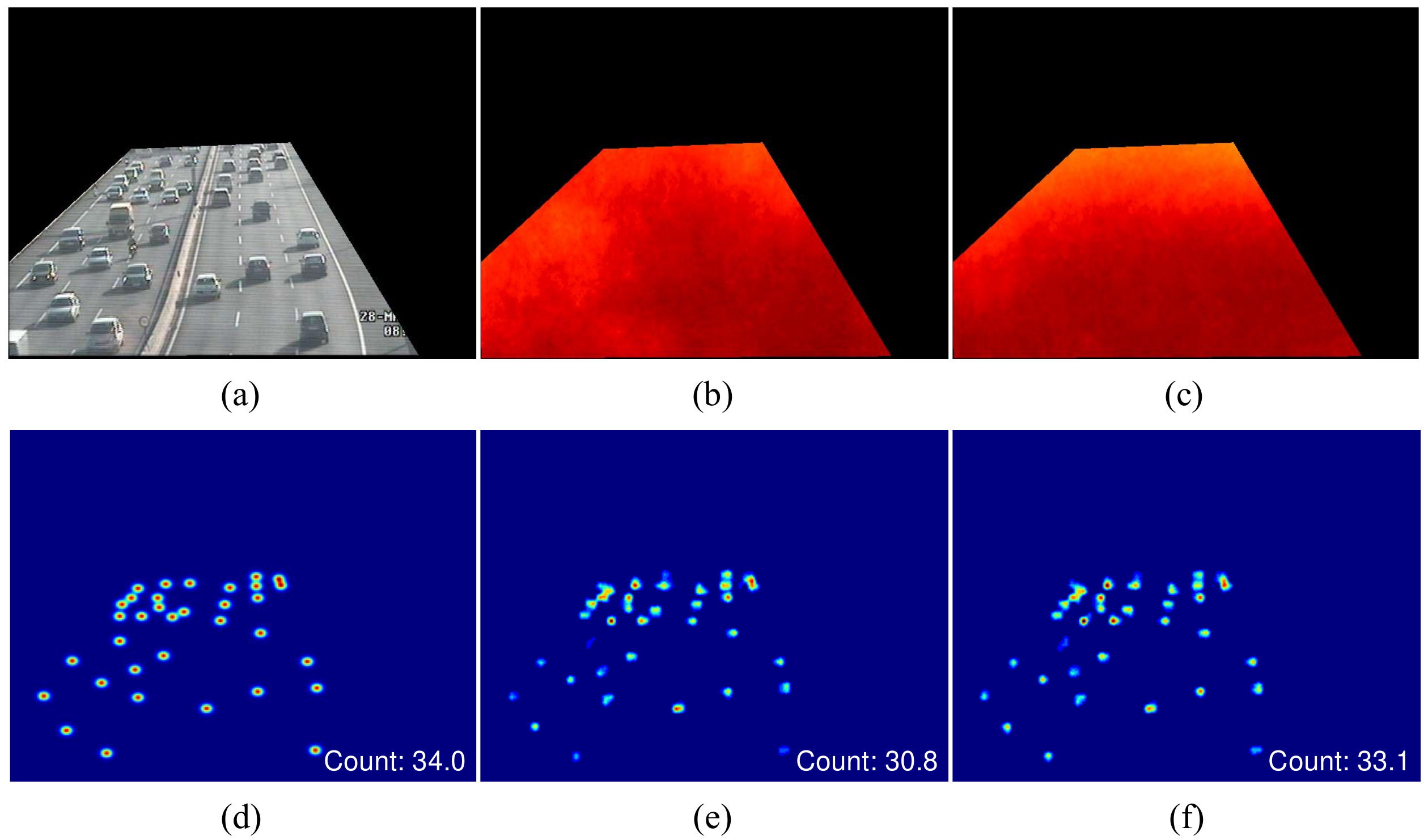}
 \scriptsize
\caption{Visualization of a test example from TRANCOS dataset. (a) is the input; (b)(c) / (e)(f) are estimated perspective / density maps of Ours A and Ours B; and (d) is the ground truth.}
\label{fig:mul_stage}

\end{figure}

Furthermore, we emphasize that PFDNet still delivers obvious performance gain even in object counting over baseline CSRNet* and achieves $3.06$ MAE.
It is noted that perspective estimation aims to predict the depth information of a scene, which is object-unrelated.
We show a test sample in Fig.~\ref{fig:mul_stage}.
For both phases \textit{Ours A} and \textit{Ours B}, PENet generally creates fine perspective estimations, which is the guarantee of fine performance.
It is generally known that the perspective map is roughly of horizontal-linear-growth and vertical-value-consistent for a plain scene~\cite{shi2019revisiting}.
Fig.~\ref{fig:mul_stage}(c) outperforms Fig.~\ref{fig:mul_stage}(b) in its higher accordance with the prior, and Fig.~\ref{fig:mul_stage}(f) outperforms Fig.~\ref{fig:mul_stage}(e) in predicting more accurate density maps.
These observations validate the robustness and feasibility of our method.
\\

\subsection{Comparisons with state-of-the-arts}
Among the existing deep crowd counting approaches, we select three state-of-the-art methods (\ie, LCM~\cite{hu2020count}, AMSNet~\cite{liu2020adaptive}, DM-Count~\cite{wang2020DMCount}) which show the best completing performance against our PFDNet.

(1) LCM~\cite{hu2020count} adopts NAS (Neural Architecture Search) to automatically develop a multi-scale architecture to address the scale variation in crowd counting.
However, the operations in the searching space are conventional network components, thus all these operations suffer from the intrinsic limitation in modeling continuous scale variation.

(2) AMSNet~\cite{liu2020adaptive} attempts to model continuous scales by stacking multiple manually-designed complicated and computational-heavy modules (\ie, Scale-aware Module, Mixture Regression Module and Adaptive Soft Interval Module), largely increasing the inference time (\ie~$\sim$$1.72s$ for a 1024$\times$768 image).
In contrast, our PFDNet only takes $\sim70ms$.

(3) DM-Count~\cite{wang2020DMCount} proposes optimal transport loss to replace MSE-based loss as the density map prediction constraint, so as to tackle the scale variation to a certain extent.
Nonetheless, the network (\ie, VGG-19~\cite{simonyan2014very}) adopted in DM-Count is built with conventional operators.
Therefore, continuous scales of features are compressed into discrete scales of feature representations, resulting in irreparable loss of information.
Actually, a better approach would be explicitly modeling the scale variation via our proposed PFD layer, which can be validated from the performance comparisons.

%
%

%
%

\subsection{Ablation Study}
\label{sec:ablation}

In this section, we will first demonstrate the influence of the number of PFCs stacked in our network.
Besides, we validate the feasibility of PFC as an insertable component for existing network to improve performance.
Moreover, we exploit the necessity of spatially variant receptive fields.
Furthermore, we visualize the dilation rate maps of different PFC layers.
In addition, we compare our PFC with deformable convolution to have a deeper understanding of continuous scale modeling.
Apart from these, we also verify the importance of the pre-training of PENet in our method.
Finally, we confirm the reliability of the prediction of PENet.

\subsubsection{Influence of the Number of PFCs}
\label{sec:number_of_persp_convs}
\begin{figure}[!t]
\centering
\includegraphics[width=0.47\textwidth]{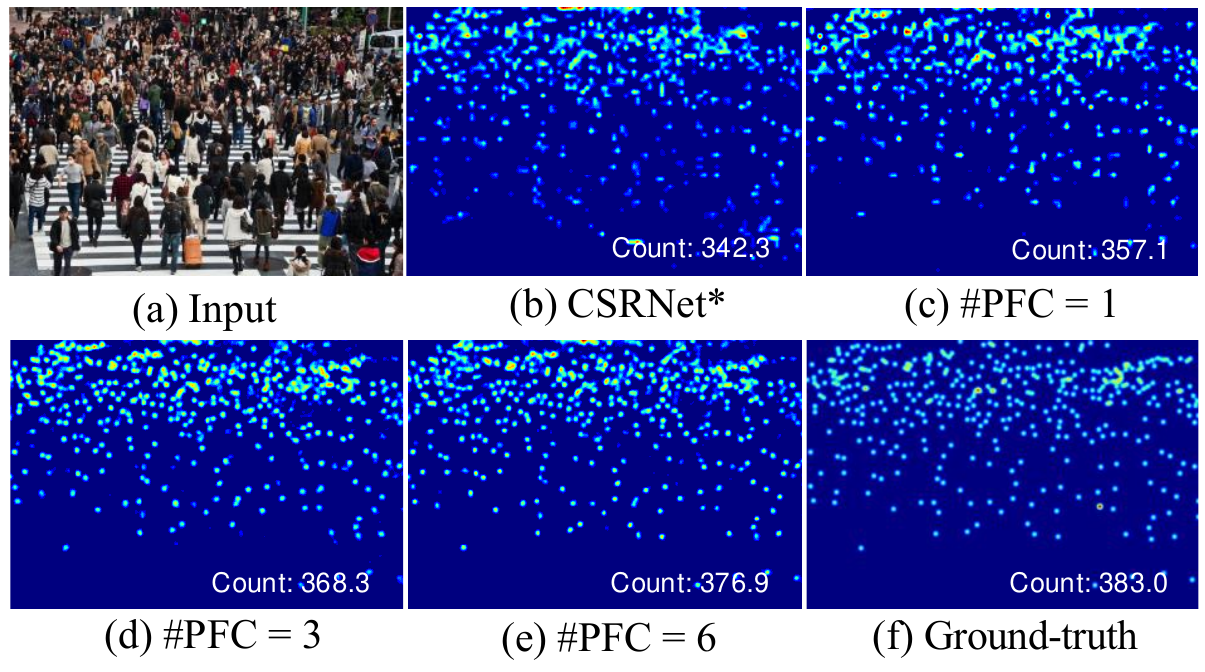}
\scriptsize
\caption{Density maps predicted by replacing different numbers of dilated convolutions with our PFCs. CSRNet* is the baseline and is equivalent to the situation of `\#PFC = 0'.}
\label{fig:mul_pfc}
\end{figure}

Table~\ref{table:differnet_modules}
shows the performance of our network when replacing dilated convolutions in CSRNet* with our proposed PFCs, as Fig.~\ref{fig:PFDNet} shows.
We know that CSRNet* is built upon VGG16-BN backbone with six dilated convolutions stacked afterward, and finally ends with a normal convolution to output density map prediction.
We validate the effectiveness of our method by replacing dilated convolution in CSRNet* with PFC layer by layer, in the order from top to bottom.
The performance increases with the number of applied PFCs, from 65.1 and 9.8 MAE as the baseline to peak values 53.8 and 6.5 MAE with six PFCs on Part A and Part B, respectively.
Fig.~\ref{fig:mul_pfc} demonstrates the predicted density maps when applied several PFCs.
It is seen that density maps are gradually turning to be more accurate on either small scales or large ones.

Here, we need to make two clarifications.

i) Better performance may be achieved when extra PFCs are applied and fine parameter-tuning is adopted, however, exploring additional PFCs is unfair in comparison with the original CSRNet*.
Therefore, we do not exploit more PFCs.

ii) Intuitively, our PFC is also able to substitute normal convolution layer which can be regarded as a special type of dilated convolution with dilation rate $1$.
Whereas, when we replace all convolutions in CSRNet* with our PFCs, we do not obtain continuous performance improvements over different datasets.
To be precise, performances of ShanghaiTech A / B turn out to be $55.4$ / $7.1$ MAE.
There exist two main reasons for the degradation.
Firstly, we do not conduct experiments on substituting normal convolutions layer by layer, which requires massive computation.
Thus, the optimal number of normal convolutions to be replaced with PFCs are not fully searched.
Secondly, we do not train our PFDNet in an incremental scheme, as it is quite complicated, though such training strategy usually leads to better performance.
Instead, the parameters in PFCs are initialized with the same setting and backbone is initialized with VGG16-BN as described in Sec.~\ref{sec:training_details}.
Based on these observations, we construct our final network as the architecture that all six dilated convolutions in CSRNet* are replaced with our proposed PFCs, resulting in $70$ ms total time cost for a $1024 \times 768$ image in testing.
%

%

%
%

\begin{table}[!ht]
   \begin{center}
      \scriptsize
	 \caption{Influence of the number of stacked PFCs in our method on ShanghaiTech dataset.}
	 \label{table:differnet_modules}
    \resizebox{0.8\hsize}{!}{
      \begin{tabular}{ c | c c |c c}
      \toprule
      \multirow{2}{*}{\# of PFCs} & \multicolumn{2}{c|}{Part A} & \multicolumn{2}{c}{Part B} \\
      \cline{2-5}
      & MAE & RMSE & MAE & RMSE \\
      \hline
      0 & 65.1 & 112.4 & 9.8 & 15.2 \\
      1 & 63.3 & 106.6 & 8.6 & 13.9 \\
      2 & 60.4 & 100.3 & 7.9 & 13.3 \\
      3 & 57.8 & 96.8 & 7.5 & 12.4 \\
      4 & 56.2 & 92.1 & 7.1 & 11.5 \\
      5 & 54.4 & 90.5 & 6.7 & 11.1 \\
      6 & \textbf{53.8} & \textbf{89.2} & \textbf{6.5} & \textbf{10.7} \\
      \bottomrule
  	\end{tabular}
  	}
  	\end{center}

\end{table}

%
%

\subsubsection{Extensibility to Another Backbone}
\label{ResNet_with_PFC_blocks}
In order to verify the extensibility of our PFC, we build a new baseline that consists of a truncated ResNet-50~\cite{he2016deep} with the first $22$ convolutional layers, four dilated convolutions (with dilation rate $2$ and $512$, $256$, $128$, $64$ channels respectively)  and a normal convolution that aims to reduce the channel dimension to $1$.
Table~\ref{table:res_pfc}
shows the comparison results on ShanghaiTech Part A / B, where \emph{ResNet-50 (backbone)} is the backbone itself, and \emph{ResNet-50 (partly PFC)} only replaces the four dilated convolutions with our PFCs and \emph{ResNet-50 (fully PFC)} replaces all convolutions with PFCs.
All models are trained for $300$ epochs with the first $22$ convolutional layers initialized by pre-trained weights on ImageNet~\cite{deng2009imagenet}.
For \emph{ResNet-50 (backbone)}, we get $89.5 / 18.6$ MAE on ShanghaiTech Part A / B.
When we replace all dilated convolutions with PFCs, a significant performance gain of $13.1 / 5.9$ MAE on ShanghaiTech Part A / B.
Besides, if all normal convolutions are also be replaced by PFCs, we observe further gain of $2.3 / 1.5$ MAE.
This indicates that PFC is not only a superior alternative to dilation convolution and even has the potential to substitute normal convolution.
Such observations validate the effectiveness and extensibility of the PFC on another backbone.
\begin{table}[!ht]
   \begin{center}
     \scriptsize
    \caption{Comparisons on the ResNet-50 backbone.}
\label{table:res_pfc}
    \resizebox{0.95\hsize}{!}{
      \begin{tabular}{ c | c  c | c  c}
      \toprule
      \multirow{2}{*}{Method} & \multicolumn{2}{c|}{Part A} & \multicolumn{2}{c}{Part B} \\
       & MAE & RMSE & MAE & RMSE \\
     \hline
      ResNet-50 (backbone) & 89.5 & 163.2 & 18.6 & 26.6 \\
      ResNet-50 (partly PFC) & 76.4 & 145.3 & 12.7 & 19.2 \\
      ResNet-50 (fully PFC) & \textbf{74.1} & \textbf{140.5} & \textbf{11.2} & \textbf{16.8} \\
    \bottomrule
  	\end{tabular}
  	}
  	\end{center}

\end{table}

\subsubsection{Necessity of Spatially Variant Receptive Fields}
\label{sec:ness_receptive_fields}
In this section, we conduct an experiment to validate the vital importance of spatially variant dilation rate guided by perspective maps.
Specifically, we average the perspective map of each image, and use the mean perspective maps to replace the original ones.
Then, we train our PFDNet with the mean perspective maps and name the trained model as \textit{Ours (mean perspective map)}.
In Table~\ref{table:different_rate_maps},
\textit{Ours (mean perspective map)} performs much worse than \textit{Ours (original perspective map)} does.
It verifies that adaptive receptive allocation is essential for estimating accurate pedestrian counts.
Although the severe performance degradation it suffers when we adopt mean perspective map for each image, it still slightly outperforms CSRNet*.
In fact, \textit{Ours (mean perspective map)} can be taken as a weaken version of PFDNet, in which dynamic receptive fields allocation is only performed among images instead of among pixels.

\begin{table}[!ht]
   \begin{center}
  \scriptsize
\caption{Comparisons on different dilation rate maps.}
\label{table:different_rate_maps}
    \resizebox{0.95\hsize}{!}{
      \begin{tabular}{ c | c  c | c  c}
      \toprule
      \multirow{2}{*}{Method} & \multicolumn{2}{c|}{Part A} & \multicolumn{2}{c}{Part B} \\
      \cline{2-5}
       & MAE & RMSE & MAE & RMSE \\
     \hline
      Ours (original perspective map) & \textbf{53.8} & \textbf{89.2} & \textbf{6.5} & \textbf{10.7} \\
      Ours (mean perspective map) & 63.7 & 104.8 & 9.0 & 14.3 \\
      CSRNet* & 65.1 & 112.4 & 9.8 & 15.2 \\
      \bottomrule
  	\end{tabular}
  	}
  	\end{center}
\end{table}

\subsubsection{Visualization of Dilation Rate Maps}
\label{sec:visual_rate_map}
\begin{figure}[!h]
\centering
\includegraphics[width=0.5\textwidth]{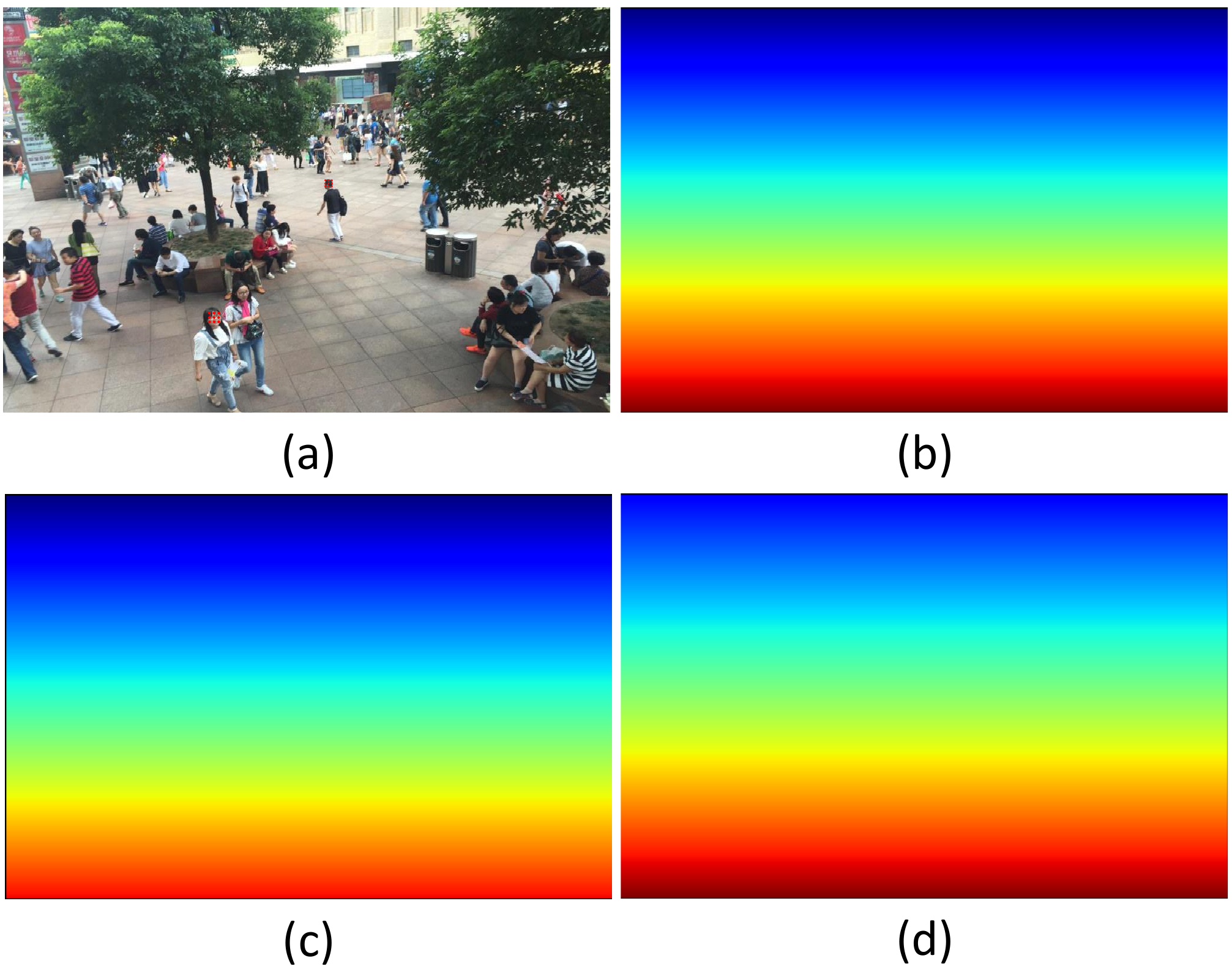}
 \scriptsize
\caption{Visualization of dilation rate maps of different PFC layers in our PFDNet. (a) is the input image, (b) is the perspective map. (c) and (d) are dilation rate maps of the first PFC (the $1$-st PFC layer, in a top to bottom order) and the last (the $6$-th PFC layer, in the same order) PFC, respectively.}
\label{fig:dilation_rate}

\end{figure}

In order to have a deeper understanding of PFC layer, we further visualize the dilation maps of different PFC layers in our PFDNet.
Fig.~\ref{fig:dilation_rate}(c) shows the dilation rate maps of the first PFC (the $1$-st PFC layer, in a top to bottom order) and Fig.~\ref{fig:dilation_rate}(d) demonstrates the last (the $6$-th PFC layer, in the same order) PFC layer in our PFDNet.
Obviously, in different PFC layers, the dilation maps are different due to the learnable parameters of $\alpha$, $\beta$, $\gamma$ and $\theta$.
Besides, for the same position, the value in Fig.~\ref{fig:dilation_rate}(c) tends to be smaller than that in Fig.~\ref{fig:dilation_rate}(d).
This accords with the fact: For the same scale of a head, when the region can be covered by a fractional dilation convolution with small dilation rate in the top layer (\ie, here is the $1$-st PFC layer in a top to bottom order), then for the lower layer (\ie, here is the $6$-th PFC layer in a top to bottom order), we need a fractional dilation convolution with larger dilation rate to cover the region.
Finally, we show the receptive fields of two heads with small red balls in (a), it can be seen that the receptive fields basically fit the scales of heads, which indicates the rationale of PFC layer.

%
%

\subsubsection{Comparison with Deformable Convolution}
\label{sec:compare_with_deform}
\begin{figure}[!h]
\centering
\includegraphics[width=0.5\textwidth]{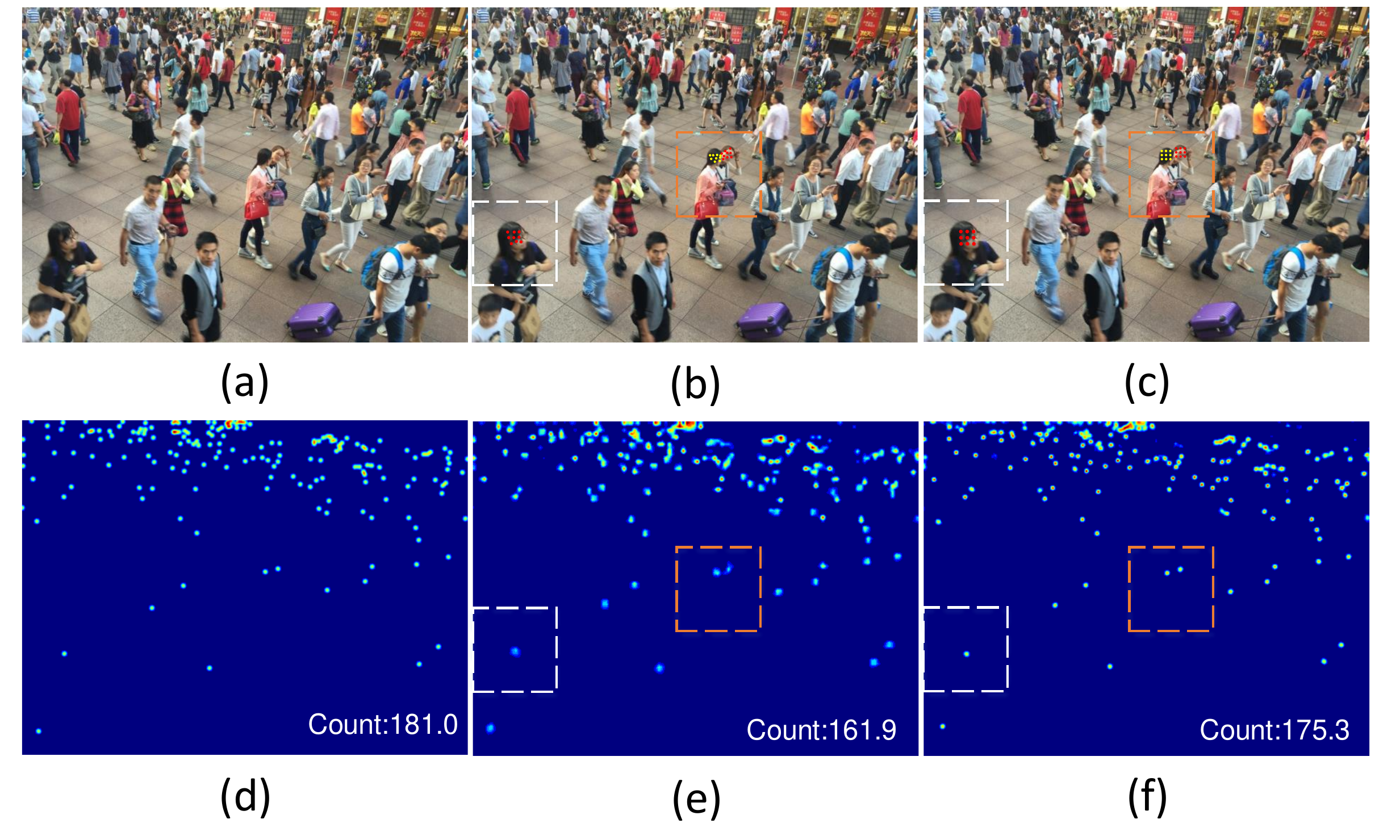}
 \scriptsize
\caption{Visualization of sampling locations of deformable convolution and PFDNet. (a) is the input image, (b) and (c) are input image with some sampling positions. (d),(e),(f) are the density maps of ground-truth, CSRNet*(DeConv) and PFDNet. It is seen that PFDNet surpasses CSRNet*(DeConv) in estimating more accurate counts and more reasonable sampling postions.}
\label{fig:deformable}

\end{figure}

Intuitively, deformable convolution~\cite{dai2017deformable} learns to allocate adaptive receptive field via enabling the offsets on the sampling positions.
Thus, we conduct an experiment by replacing normal dilated convolution in CSRNet* with deformable convolution~\cite{dai2017deformable} layer by layer, in the order from top to bottom, All the settings are kept the the same as those described in Sec.~\ref{sec:number_of_persp_convs}.
We term this network CSRNet*(DeConv).
Table~\ref{table:pfc_vs_deform} shows the performance comparison on ShanghaiTech dataset.
It is seen that when all the six dilated convolutions are replaced by deformable convolutions, CSRNet*(DeConv) delivers the best performance with MAE $56.7 / 7.6$ on ShanghaiTech A / B, respectively, however, still inferior to our PFDNet with the margin of $2.9 / 1.1$ MAE.
This indicates the superiority of our PFC against deformable convolution.
Furthermore, in Fig.~\ref{fig:deformable}, we visualize the sampling positions of PFC and deformable convolution.
Concretely, we show two typical cases.
In the region surrounded by white dashed lines, it is observed that the layout of sampling positions is approximately in the shape of a square, resulting in a roughly Gaussian-like blob in Fig.~\ref{fig:deformable}(e).
In the area marked with orange dashed lines, two heads are too close and the corresponding two groups of offsets mix with each other, leading to inadequate feature aggregation and unsatisfying prediction in Fig.~\ref{fig:deformable}(e).
While, for our PFC shown in Fig.~\ref{fig:deformable}(f), as guided by the prior of the perspective map, it can be seen that the sampling positions fall uniformly in an appropriate square in the head, leading to much more accurate and clear estimation in both cases.
%

%
%

\begin{table}[!ht]
   \begin{center}
	 \caption{Comparison of PFC with deformable convolution on ShanghaiTech dataset.}
	
    \resizebox{0.95\hsize}{!}{
    	\label{table:pfc_vs_deform}
      \begin{tabular}{ c | c c |c c}
      \toprule
      \multirow{2}{*}{\# of PFCs / DeConvs} & \multicolumn{2}{c|}{Part A} & \multicolumn{2}{c}{Part B} \\
      \cline{2-5}
      & PFC & DeConv~\cite{dai2017deformable} & PFC & DeConv~\cite{dai2017deformable} \\
      \hline
      0 & 65.1 & 65.1 & 9.8 & 9.8 \\
      1 & 63.3 & 62.6 & 8.6 & 9.2 \\
      2 & 60.4 & 60.3 & 7.9 & 8.7 \\
      3 & 57.8 & 58.4 & 7.5 & 8.3 \\
      4 & 56.2 & 57.8 & 7.1 & 8.0 \\
      5 & 54.4 & 57.1 & 6.7 & 7.7 \\
      6 & \textbf{53.8} & 56.7 & \textbf{6.5} & 7.6 \\
      \bottomrule
  	\end{tabular}
  	}
  	\end{center}

\end{table}

%
%

\subsubsection{Importance of PENet Pre-training}
\label{sec:Importance_of_pre-training}
The pre-training of PENet makes the decoder part a robust perspective map constructor.
By doing so, \textit{Ours B} can recover a roughly satisfying perspective estimation as shown in Fig.~\ref{fig:mul_stage}.
In this section, we validate the necessity the pre-training of PENet.
To this end, we conduct two experiments without pre-training of PENet on UCF\_CC\_50~\cite{idrees2013multi} and TRANCOS~\cite{guerrero2015extremely}.
As shown in Table~\ref{table:PENet_pretraining},
the results with PENet pre-training significantly outperform those without pre-training.
Such performance margin is attributed to the confusing guidance of PENet, since without pre-training, we observe that the output of PENet is still messy even after considerable epochs of training.
\begin{table}[!ht]
	\begin{center}
     \scriptsize
	\caption{The performances with / without pre-training of PENet on UCF\_CC\_50~\cite{idrees2013multi} and TRANCOS~\cite{guerrero2015extremely}.}
	
		\label{table:PENet_pretraining}
		\resizebox{0.95\hsize}{!}{
			\begin{tabular}{ c|  c c| c}
				\toprule
				\multirow{2}{*}{Method} & \multicolumn{2}{c|}{UCF\_CC\_50} &
				TRANCOS \\
				\cline{2-4}
				& MAE & RMSE & GAME 0   \\
				\hline
				PENet(w/ pre-training) & \textbf{205.8} & \textbf{289.3} & \textbf{3.06} \\
				PENet(w/o pre-training) & 264.8 & 383.5 & 5.28   \\
				\bottomrule
			\end{tabular}
		}
	\end{center}

\end{table}

\subsubsection{Reliability of PENet Prediction}
\label{sec:Reliability_of_PENet}
PENet is designed as a compromise of the situation that perspective annotations are unavailable, in which the reliability of PENet is crucial.
Therefore, we conduct an experiment to confirm the feasibility of PENet.
Table~\ref{table:reliablity_of_PENet}
demonstrates the comparisons of adopting the ground-truth or the estimated perspective map
as the guidance of PFC on ShanghaiTech Part A/B and WorldExpo'10.
MAEs are respectively $55.0$, $7.5$ and $7.3$, with a small decrease of $1.2$, $1.0$ and $0.5$, respectively.
This indicates that PENet is competent to a reasonable perspective map estimator.

\begin{table}[!ht]
	\begin{center}
	\caption{Different guidances of PFC on ShanghaiTech Part A/B and WorldExpo'10.}
		\label{table:reliablity_of_PENet}
		\resizebox{1\hsize}{!}{
			\begin{tabular}{c | c | c }
				\toprule
				Perspective Map & ShanghaiTech Part A/B & WorldExpo'10 \\
				\hline
				Estimated & 55.0/7.5 & 7.3 \\
				\hline
				Ground-truth & \textbf{53.8}/\textbf{6.5} & \textbf{6.8}  \\
				\bottomrule
			\end{tabular}
		}
	\end{center}
\end{table}

%
\section{Conclusion}
\label{sec:conclusion}
In this paper, we have presented a perspective-guided fractional-dilation convolutional network (PFDNet) for crowd counting.
The key idea of PFDNet is the perspective-guided fractional-dilation convolution, which plays a role as an insertable module that successfully handles the continuous intra-scene scale variation issue under the guidance of perspective map annotations.
We have also proposed a perspective estimation branch as well as its learning strategy, which is incorporated into our method to form an end-to-end trainable network, even without perspective map annotations.
Experiments on five benchmark datasets have shown the superiority of our PFDNet against the state-of-the-arts.

\section*{Acknowledgment}
This project is partially supported by the National Natural Scientific Foundation of China (NSFC) under Grant No.s U19A2073 and 61872118.

\ifCLASSOPTIONcaptionsoff
  \newpage
\fi

\bibliographystyle{IEEEtran}
\bibliography{IEEEabrv,egbib}

\begin{IEEEbiography}[{\includegraphics[width=1in,height=1.25in,clip,keepaspectratio]{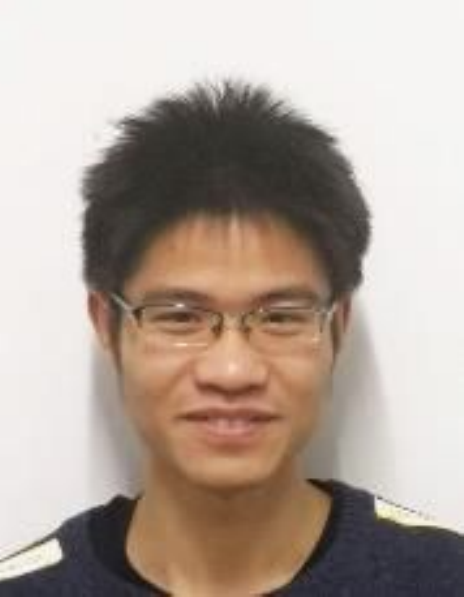}}]{Zhaoyi Yan} is a Ph.D candidate of Computer Science, in Harbin Institute of Technology, Harbin, China. His research interests include deep learning and crowd counting.
\end{IEEEbiography}

\begin{IEEEbiography}[{\includegraphics[width=1in,height=1.5in,clip,keepaspectratio]{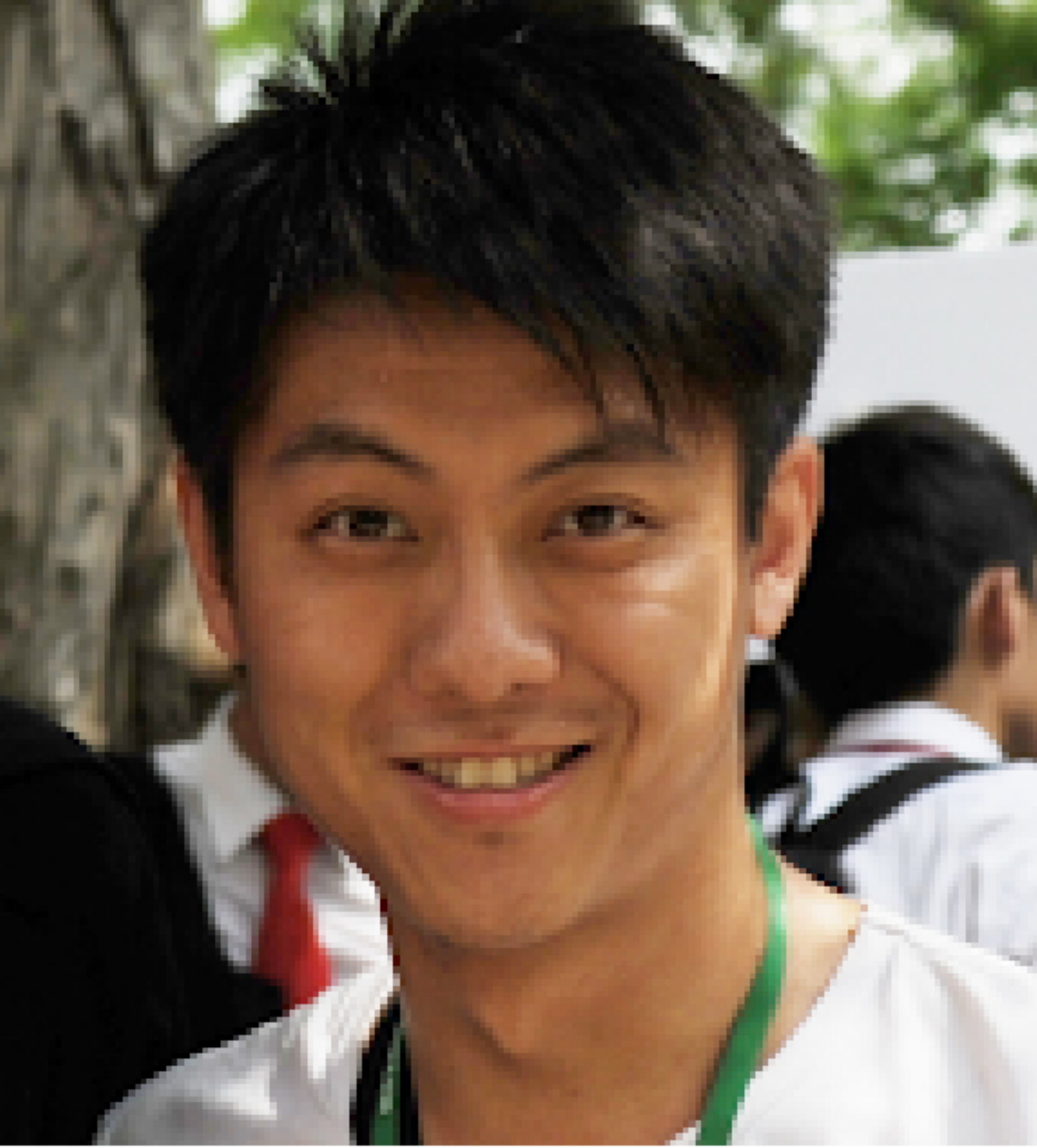}}]{Ruimao Zhang} is currently a Research Assistant Professor in the school of Data Science, The Chinese University of Hong Kong, Shenzhen (CUHK-SZ), China. He received the B.E. and Ph.D. degrees from Sun Yat-sen University, Guangzhou, China, in 2011 and 2016, respectively. He was a Postdoctoral Fellow in CUHK from 2017 to 2019. Before that, he was a visiting Ph.D. student with Hong Kong Polytechnic University (PolyU) from 2013 to 2014. His current research interests include computer vision, deep learning, and related multimedia applications.

\end{IEEEbiography}

\begin{IEEEbiography}[{\includegraphics[width=1in,height=1.25in,clip,keepaspectratio]{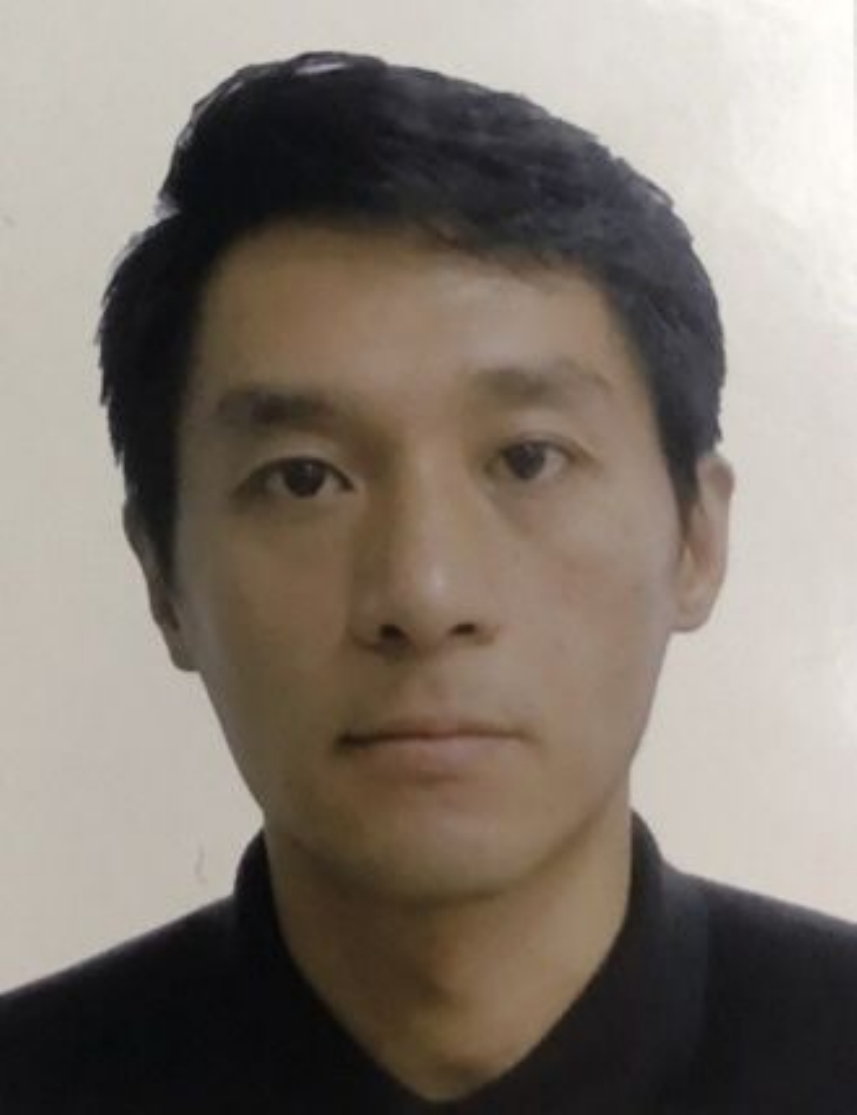}}]{Hongzhi Zhang} received his Ph.D. degree in computer science and technology from Harbin institute of Technology (HIT), China, in 2007. He is an associate Professor at the School of Computer Science and Technology, HIT. His research interests include theoretic approaches to problems in signal processing, computer vision and machine learning. His research has been supported by grants from the National Natural Science Foundation of China.
\end{IEEEbiography}

\begin{IEEEbiography}[{\includegraphics[width=1in,height=1.25in,clip,keepaspectratio]{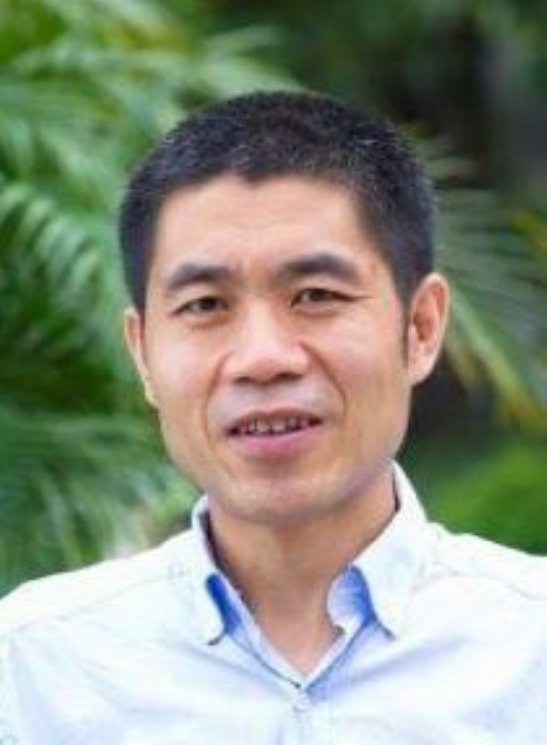}}]
{Qingfu Zhang} (M'01-SM'06-F'17) received the
BSc degree in mathematics from Shanxi University, China in 1984, the MSc degree in applied mathematics and the PhD degree in information engineering
from Xidian University, China, in 1991 and 1994, respectively. He is a Chair Professor of Computational Intelligence at the Department of Computer Science, City University of Hong Kong. His main research interests include evolutionary computation, optimization, neural networks, data analysis, and
their applications. Dr. Zhang is an Associate Editor of the IEEE Transactions on Evolutionary Computation and the IEEE
Transactions on Cybernetics. He is a Clarivate Highly Cited Researcher in Computer Science for four consecutive years from 2016.
\end{IEEEbiography}

\begin{IEEEbiography}[{\includegraphics[width=1in,height=1.25in,clip,keepaspectratio]{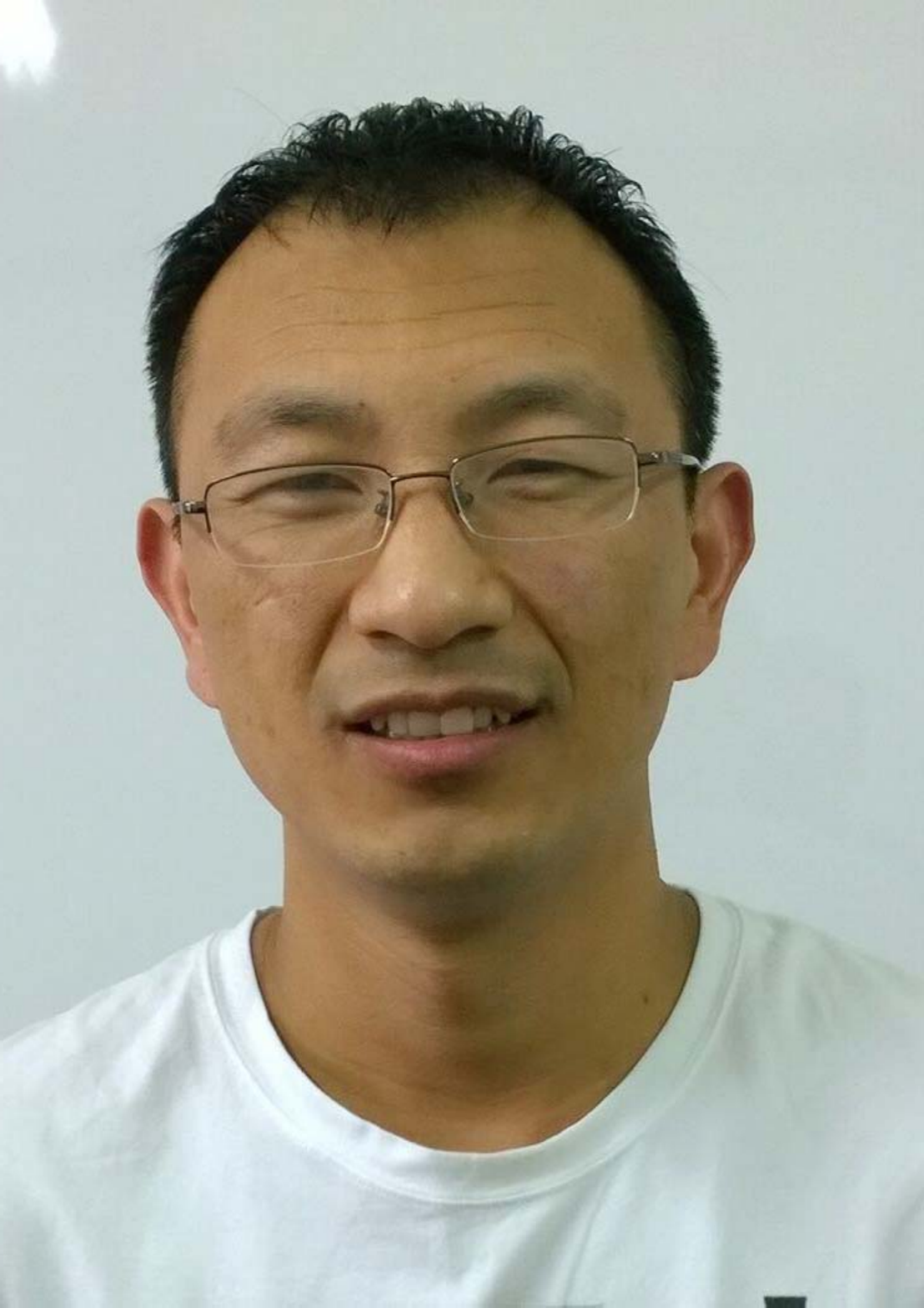}}]{Wangmeng Zuo}(M'09-SM'14)
 received the Ph.D. degree in computer application technology from the Harbin Institute of Technology, Harbin, China, in 2007.\underline{}
 He is currently a Professor in the School of Computer Science and Technology, Harbin Institute of Technology. His current research interests include image enhancement and restoration, image and face editing, object detection, visual tracking, and image classification. He has published over 100 papers in top tier academic journals and conferences. He has served as a Tutorial Organizer in ECCV 2016, an Associate Editor of the \emph{IET Biometrics} and \emph{Journal of Electronic Imaging}.
\end{IEEEbiography}

\end{document}